\documentclass[journal]{IEEEtran}

\usepackage{cite}
\usepackage{romannum}
\usepackage{amsmath,amsfonts,amssymb,enumerate,cases}
\usepackage{amsthm}
\usepackage{subfigure}
\usepackage{caption}
\usepackage[final]{lscape,graphicx,epsfig}
\usepackage{graphicx}
\usepackage{color}
\usepackage[table]{xcolor}

\usepackage{bigints}
\usepackage{relsize}
\usepackage{changepage}

\usepackage{algorithm}
\usepackage[noend]{algpseudocode}
\usepackage{algpseudocode}
\makeatletter
\def\BState{\State\hskip-\ALG@thistlm}
\makeatother

\graphicspath{ {image/} }

\usepackage[english]{babel}

\usepackage{graphicx}

\newcounter{definition}
\newenvironment{definition}[1][]{\refstepcounter{definition}\par\medskip\noindent%
	\textit{Definition~\thedefinition. #1} \rmfamily}{\medskip}

\begin{document}
	
\title{Coverage-Based Designs Improve Sample Mining and Hyper-Parameter Optimization}


\author{Gowtham Muniraju$^{\dagger}$, Bhavya Kailkhura$^{\ddagger}$, Jayaraman J. Thiagarajan$^{\ddagger}$, Peer-Timo Bremer$^{\ddagger}$,\\
	 Cihan Tepedelenlioglu$^{\dagger}$, \textit{Senior Member, IEEE} and Andreas Spanias$^{\dagger}$, \textit{Fellow IEEE}.
	\thanks{$^{\dagger}$ G. Muniraju, C. Tepedelenlioglu and A. Spanias are with School of
		ECEE, Arizona State University. Email: \{gmuniraj, cihan, spanias\}@asu.edu.
			
		$^{\ddagger}$ B. Kailkhura, J. J. Thiagarajan and P-T. Bremer are with the Center for Applied Scientific Computing, Lawrence Livermore National Laboratory. Email: \{kailkhura1, jjayaram, bremer5\}@llnl.gov}
	}

\markboth{Improving Hyper-Parameter Optimization using Coverage-Based Sample Designs}%
{Improving Hyper-Parameter Optimization using Coverage-Based Sample Designs}

\maketitle

\begin{abstract}
	Sampling one or more effective solutions from large search spaces is a recurring idea in machine learning, and sequential optimization has become a popular solution. Typical examples include data summarization, sample mining for predictive modeling and hyper-parameter optimization. Existing solutions attempt to adaptively trade-off between global exploration and local exploitation, wherein the initial exploratory sample is critical to their success. While discrepancy-based samples have become the \textit{de facto} approach for exploration, results from computer graphics suggest that coverage-based designs, e.g. Poisson disk sampling, can be a superior alternative. 
	In order to successfully adopt coverage-based sample designs to ML applications, which were originally developed for $2-d$ image analysis, we propose fundamental advances by constructing a parameterized family of designs with provably improved coverage characteristics, and by developing algorithms for effective sample synthesis. 
	Using experiments in sample mining and hyper-parameter optimization for supervised learning, we show that our approach consistently outperforms existing exploratory sampling methods in both blind exploration, and sequential search with Bayesian optimization.
\end{abstract}

\IEEEpeerreviewmaketitle

\section{Introduction}

\subsection{Sampling in Machine Learning}
Sample design has been a long-standing research area in statistics~\cite{fisher}, and has now become a crucial problem in machine learning and AI, particularly with the emergence of numerous data-driven learning paradigms. The notion of sampling appears in a variety of contexts, ranging from summarizing complex data~\cite{bock2012analysis_intro1}, generating mini-batches for effective neural network training~\cite{csiba2018importance_intro2}, metric learning~\cite{wu2017sampling} to hyper-parameter search~\cite{bergstra2012random,bousquet2017critical}, reinforcement learning~\cite{sutton2018reinforcement_intro4,asmuth2009bayesian_intro3} and knowledge transfer~\cite{milli2017interpretable}. A common goal in these seemingly diverse applications is to identify one or more effective solutions from a large search space, using the smallest amount of resources. In principle, there are two competing strategies while performing sampling~\cite{gupta2006interplay}: \textit{exploitation}, which probes a limited region in the search space with the hope of improving an already identified solution; and \textit{exploration}, which probes a larger part of the search space with the hope of finding solutions that are yet to be refined. In practice, sequential sampling methods that can trade-off between exploration and exploitation strategies are preferred~\cite{doucet2000sequential}. However, given the large volume of typical search spaces and restrictions on resources (time and compute), the exploration step is highly critical to reduce the uncertainties to an extent that the exploitation step can be expected to succeed. Over the last several decades, a large class of exploratory sampling techniques have been developed~\cite{Ebeida:2012,Ebeida2014,Sampling:survey}. Though the overarching objective is to cover the search space uniformly, it is well known that uniformity alone does not suffice. For example, optimal sphere packings lead to highly uniform designs, yet are prone to causing aliasing artifacts. Consequently, effective exploration requires to balance uniformity and randomness in the search space, often evaluated using heuristic measures such as discrepancy~\cite{niederreiter1992random}. More recently, the \textit{pair correlation function} (PCF) has been found to be a more useful statistic for evaluating the quality of sample designs~\cite{Oztireli:2012,Kailkhura:2016,Kailkhura:2016:SBN,kailkhura2018spectral}.

While discrepancy-based quasi-random designs have been commonly utilized in several applications~\cite{bergstra2012random,bergstra2011algorithms}, the computer graphics community has had long-standing success with coverage-based designs, in particular Poisson Disk Sampling (PDS)~\cite{Dippe:1985,Cook:1986}. The works in~\cite{Dippe:1985,Cook:1986} were the first to introduce PDS for turning regular aliasing patterns into featureless noise, which makes them perceptually less visible. Their works were inspired by the seminal work of Yellott et.al.~\cite{Yellott:1983}, who observed that the photo-receptors in the retina of monkeys and humans are distributed according to a Poisson disk distribution. For the first time in~\cite{Kailkhura:2016}, PDS was formally defined using the pair correlation function and used to obtain theoretical bounds on achievable coverage properties. Despite their well-established success in image/volume rendering~\cite{Ebeida:2011,Ebeida:2012,Heck:2013}, coverage-based designs have not been adopted in the machine learning community. Recently, in~\cite{kailkhura2018spectral}, Kailkhura \textit{et al.} developed a generic spectral sampling framework, that encompasses several existing designs including PDS, blue noise~\cite{Heck:2013} and variants~\cite{Kailkhura:2016:SBN}, by jointly analyzing the spatial and spectral properties of sample distributions. Though this framework enjoys several desirable properties in theory, constructing an optimal design and actually synthesizing samples that match these characteristics are challenging. When designed sub-optimally, a spectral sample can perform worse than other random sampling strategies. In addition, the sample synthesis is based solely on PCF matching, which is a summary $1-$D statistic of high-dimensional point clouds, thus making this optimization very challenging in practice.

\subsection{Proposed Work}
In this work, we propose to develop novel coverage-based designs for challenging machine learning problems, namely sample mining in predictive modeling and hyper-parameter optimization. Building upon the theoretical foundations from~\cite{kailkhura2018spectral}, we argue that larger coverage is critical to improving the expected performance of sample designs in ML tasks that require the exploration of complex optimization surfaces. Further, we make the following key contributions to produce highly effective samples in practice:
\begin{itemize}
\item We introduce a new parameterized family of coverage-based designs using the pair correlation function, which generalizes existing constructions such as~\cite{Kailkhura:2016, kailkhura2018spectral}, for machine learning applications;
\item Using tools from spectral sampling theory, we show that the proposed sample design achieves the largest coverage so far;
\item For the first time, we develop an efficient strategy to find the ``optimal'' parameters of a design (i.e. with largest coverage) for a given sample size and dimensionality;
\item We design a scalable and effective sample synthesis algorithm that consistently outperforms existing PCF matching approaches such as~\cite{Oztireli:2012} and~\cite{kailkhura2018spectral}.
\item Using empirical studies on predictive modeling, we demonstrate that the proposed sample design outperforms existing discrepancy-based and coverage-based designs, particularly under low sampling rates.
\end{itemize}
The proposed coverage-based design is based on systematically trading-off randomness characteristics of a point distribution with coverage to enable improved performance. Such a controlled random sampling is mathematically represented using the PCF and analyzed via the spectral sampling principles from~\cite{kailkhura2018spectral}. Surprisingly, we find that the achievable coverage of the proposed design is significantly larger than a conventional PDS design ($\sim 25\%-40\%$ increase for the same configuration). Further, our synthesis algorithm consistently produces high-quality samples and is highly robust, as evidenced by the performance variance across multiple realizations. 

In order to demonstrate the importance of coverage-based designs in challenging applications, we consider the problem of hyper-parameter tuning while building ML models. To this end, we consider scenarios where we rely solely on exploration (\textit{blind sampling}), similar to~\cite{bergstra2012random}, and where we use the exploratory samples to initialize a Bayesian optimization pipeline with expected improvement as the acquisition function, as carried out in~\cite{bergstra2011algorithms}. We perform empirical studies with (a) a standard feature extractor-classifier pipeline, and (b) deep neural networks that perform end-to-end learning. Our results show that the proposed sample design consistently outperforms state-of-the art exploratory sampling methods including Latin Hyper Cube (LHS), Quasi-Monte Carlo (QMC) designs~\cite{niederreiter1992random} and spectral samples in~\cite{kailkhura2018spectral}. Interestingly, we observed significant improvements even in the Bayesian  optimization cases, which clearly emphasizes the importance of the initial exploration step. In summary, the effectiveness of our approach even with small sample sizes establishes it as a powerful exploratory sampling technique for ML/AI applications.

\section{Coverage-Based Sample Designs}
\label{back}

Though a variety of discrepancy measures are commonly used for exploratory sampling, our focus is on coverage-based designs. In this section, we briefly describe the mathematical tools required for the design and analysis of coverage-based sampling. Subsequently, we discuss two popular coverage-based designs from~\cite{Kailkhura:2016} and~\cite{kailkhura2018spectral} respectively.

Broadly speaking, a reasonable objective for exploratory sampling is to ensure that the samples are random, thus providing an equal chance of finding meaningful solutions anywhere in the search space. However, in order to ensure diversity, a second objective is often considered, which is to cover the space uniformly. In this paper, we consider the general class of coverage-based sample designs~\cite{kailkhura2018spectral}:
\begin{definition}(\textit{Coverage-based Design})
\label{def:design}
	A set of $N$ random samples $\{\mathbf{X_i}\}_{i=1}^N$ in a search space $\mathcal{D}$ can be characterized as a coverage-based design, if $\{\mathbf{X_i} = \mathbf{x_i}\in \mathcal{D};\;i=1,\cdots N\}$ satisfy the following two objectives:
	\begin{itemize}
		\item $\forall \mathbf{X_i},\;\forall \triangle \mathcal{D} \subseteq \mathcal{D}\;:\;Pr(\mathbf{X_i}=\mathbf{x_i}\in \triangle \mathcal{D})=\frac{1}{\triangle \mathcal{D}} \int_{\triangle \mathcal{D}}\mathbf{dx}$; \\
		\item $\forall \mathbf{x_i},\mathbf{x_j}\;:\;||\mathbf{x_i}-\mathbf{x_j}||\geq r_{\text{min}}$,
	\end{itemize}
\end{definition}where $r_{\text{min}}$ is referred to as the \textit{coverage radius} (or disk size). In this definition, the first objective states that the probability of a random sample $\mathbf{X_i}$  falling inside a subset $\triangle \mathcal{D}$ of $\mathcal{D}$ is equal to the hyper-volume of $\triangle \mathcal{D}$. The
second condition enforces the disk constraint for improving coverage. Since, there existed no quality metrics to jointly characterize the coverage and randomness properties, several recent works have adopted the pair correlation function~\cite{Oztireli:2012} as a quality metric.
\begin{definition}(\textit{Pair Correlation Function})
	\label{def:pcf}
	Let us denote the intensity of a point process $\mathcal{X}$ as
	$\lambda(\mathcal{X})$, \textit{i.e.}, the average number of points in an
	infinitesimal volume around $\mathcal{X}$. For isotropic point
	processes, this is a constant. To define the product density
	$\beta$, let $\{B_i\}$ denote the set of infinitesimal spheres
	around the points, and $\{dV_i\}$ indicate the volume measures of
	$B_i$.  Then, we have $Pr(\mathbf{X_1} = \mathbf{x_1}, \cdots, \mathbf{X_N}= \mathbf{x_N}) = \beta(\mathbf{x_1}, \cdots,
	\mathbf{x_N})dV_1\cdots dV_N$ which represents the probability of having points $\mathbf{x_i}$ in $\{B_i\}$.
	In the isotropic case, for a pair of points, $\beta$ depends only on
	the distance between them, and hence
  $\beta(\mathbf{x_i},\mathbf{x_j})=\beta(||\mathbf{x_i}-\mathbf{x_j}||)=\beta(r)$
	and $Pr(r)=\beta(r)dV_idV_j$. The PCF is then defined as $G(r)={\beta}/{\lambda^2}$.
\end{definition}

\begin{figure}[t]
	\centering
	\includegraphics[clip=true, width=0.8\linewidth]{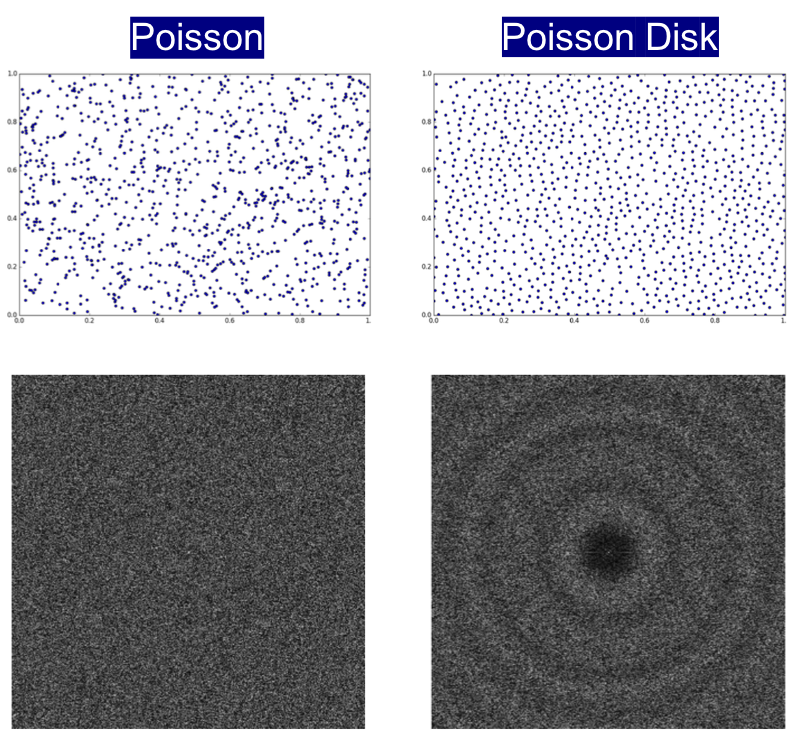}
	\caption{Illustration of {$2$-d} patterns obtained using Poisson and Poisson disk sampling. We show the point distribution (top) and the power spectral density (bottom) for each case.}
	\label{fig:pdpds}
	\vspace{-0.1in}
\end{figure}

Alternatively, Fourier analysis can be utilized for understanding the qualitative properties of sampling patterns. For isotropic samples, a metric of interest is the radially-averaged power spectral density (PSD), which describes how the signal power is distributed over spatial frequencies.

\begin{definition} (\textit{Radially-averaged PSD})
	{
		For a finite set of $N$ points, $\{\mathbf{x}_j\}_{j=1}^N$, in a
		region with unit volume, the PSD of the sampling function $\sum_{j=1}^N\delta(\mathbf{x}-\mathbf{x}_j)$ is defined as
		\begin{equation}
		P(\mathbf{k}) = \frac{1}{N} |S(\mathbf{k})|^2 = \frac{1}{N} \sum_{j,\ell} e^{-2\pi i \mathbf{k}.(\mathbf{x}_{\ell} - \mathbf{x}_j)},  
		\end{equation}
		where $|.|$ denotes the $\ell_2$-norm and $S(\mathbf{k})$ denotes the Fourier transform of the sampling function}.
\end{definition}

Interestingly, there is a well-defined connection between the PCF of a sample design and its radially-averaged PSD, and this connection is central to the proposed work.

\begin{definition} (\textit{Linking PCF and PSD})
	For an isotropic sample design with $N$ points,
	$\{\mathbf{x}_j\}_{j=1}^N$, in a $d$-dimensional region, the radially averaged
	power spectral density $P(k)$ and the pair correlation function $G(r)$ are related as follows:
	\begin{equation}
	\label{eq:link}
	P({k})= 1+\dfrac{N}{V} (2\pi)^{\frac{d}{2}}k^{1-\frac{d}{2}}H_{\frac{d}{2}-1}\left[r^{\frac{d}{2}-1}G(r)-1\right],
	\end{equation}
	where $k$ is the frequency index, $V$ is the volume of the sampling region and $H_d[.]$ denotes the Hankel transform,
	\begin{equation*}
	H_d(f(r))(k)=\int_{0}^{\infty}r J_d(kr)f(r) dr,
	\end{equation*}
	with $J_d(.)$ denoting the Bessel function of order $d$.
\end{definition}

Finally, it is important to note that, not every PCF construction is physically realizable by a sample design. In fact, there are two necessary mathematical conditions~\footnote{Whether or not these two conditions
	are not only necessary but also sufficient is still an open question
	(however, no counterexamples are known).} that a design must
satisfy to be realizable.

\begin{definition}(\textit{Realizability})
	\label{real}
	A PCF can be considered to be {potentially} realizable through a sample design, if it satisfies:
	\begin{itemize}
		\item the PCF must be non-negative, i.e., $G(r)\geq 0,\;\forall r$, and
		\item the corresponding PSD must be non-negative, i.e., $P(k)\geq 0,\;\forall k$.
	\end{itemize}
\end{definition}

\subsection{Poisson Disk Sampling}
The well-known Poisson design (Figure \ref{fig:pdpds}(a)) enforces only the first condition from Definition \ref{def:design}, in which case the number of samples that fall inside any subset $\triangle \mathcal{D} \subseteq \mathcal{D}$ obeys a discrete Poisson distribution. Consequently, Poisson disk sampling~\cite{Kailkhura:2016} (Figure \ref{fig:pdpds}(b)) that explicitly enforces the disk constraint is considered to be optimal in this context. Several widely adopted strategies for generating Poisson disk samples rely on the heuristic idea of dart throwing~\cite{Dippe:1985,Cook:1986,Ebeida:2012,Ebeida2014}, which uses as many darts as required to cover the search space, while not violating the disk criterion. Despite its effectiveness, its primary shortcoming is the choice of termination condition, since it is not easy to quantify the coverage and randomness properties. This motivated the use of pair correlation function (PCF)~\cite{Oztireli:2012} to summarize spatial characteristics of a sampling pattern, using which Kailkhura \textit{et al.}~\cite{Kailkhura:2016} formally defined Poisson disk sampling for the first time (Figure \ref{fig:pcf_coverage}(a)).

\begin{definition}(\textit{Poisson disk sampling})~\cite{Kailkhura:2016} Given the desired disk size $r_{min}$, PDS is defined using the PCF as
\begin{equation}
G(r-r_{\text{min}}) = \left\{ \begin{array}{rll}
0  & \mbox{if}\ r< r_{\text{min}} \\
1  & \mbox{if}\ r \geq r_{\text{min}}.
\end{array}\right.
\label{eq:pds}
\end{equation}	
\end{definition}Note that, the disk radius $r_{\text{min}}$ is referred as the coverage.

\subsection{Space-Filling Spectral Design}
\begin{figure*}[t]
	\centering
	\subfigure[Poisson Disk Sampling~\cite{Kailkhura:2016}]{\includegraphics[clip=true, width=0.32\linewidth]{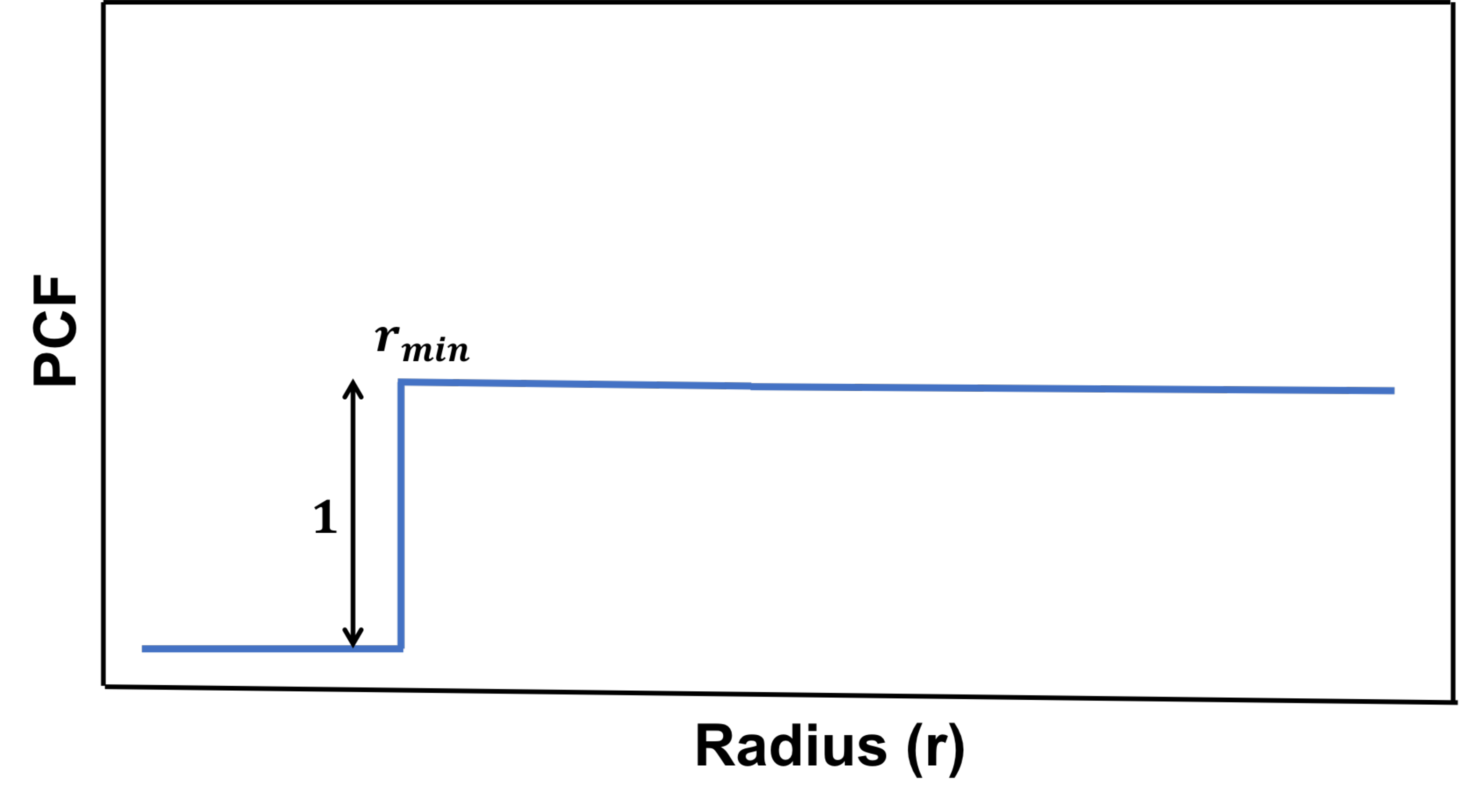}}
	\subfigure[Space Filling Spectral Design~\cite{kailkhura2018spectral}]{\includegraphics[clip=true, width=0.32\linewidth]{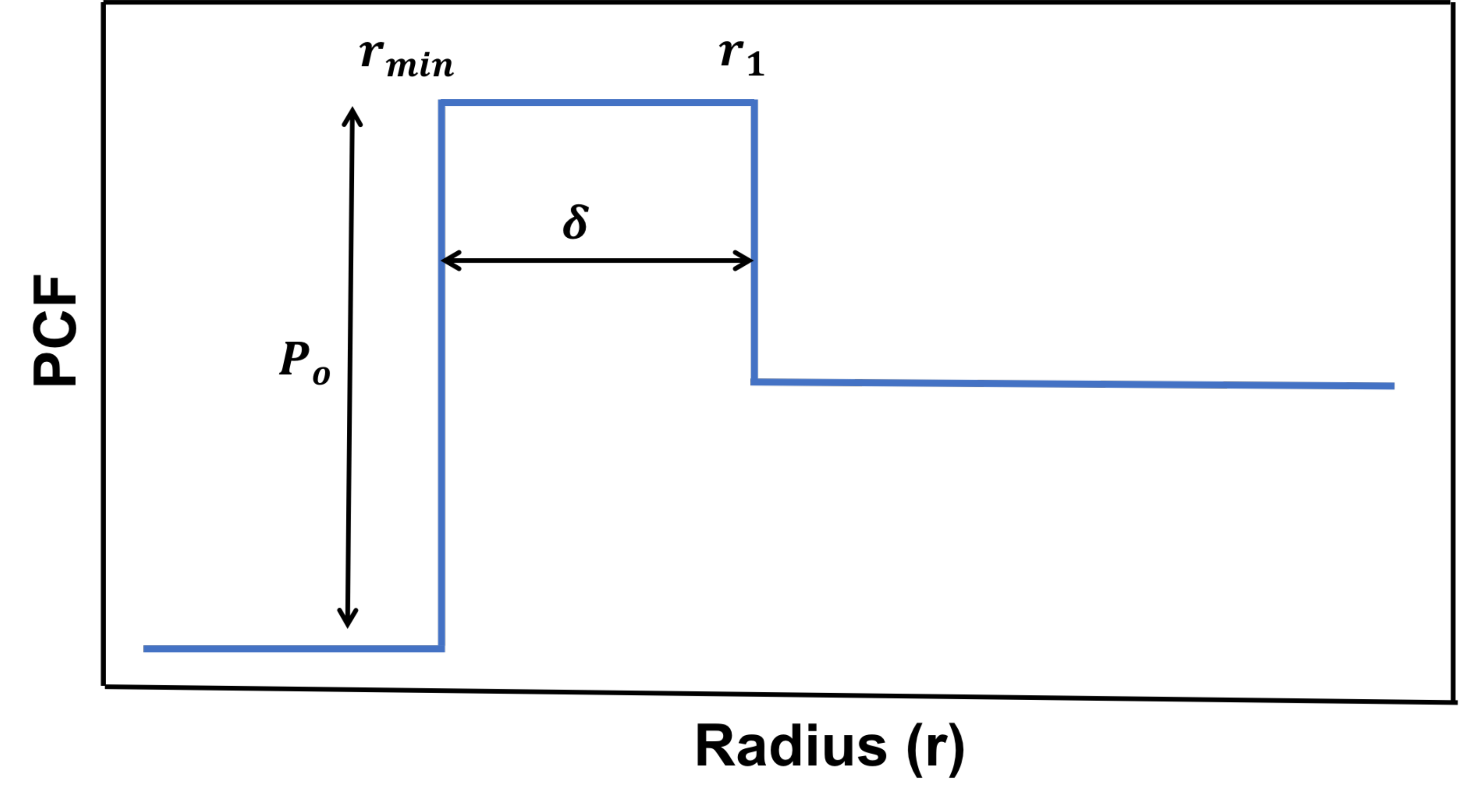}}
	\subfigure[Proposed Design]{\includegraphics[clip=true, width=0.32\linewidth]{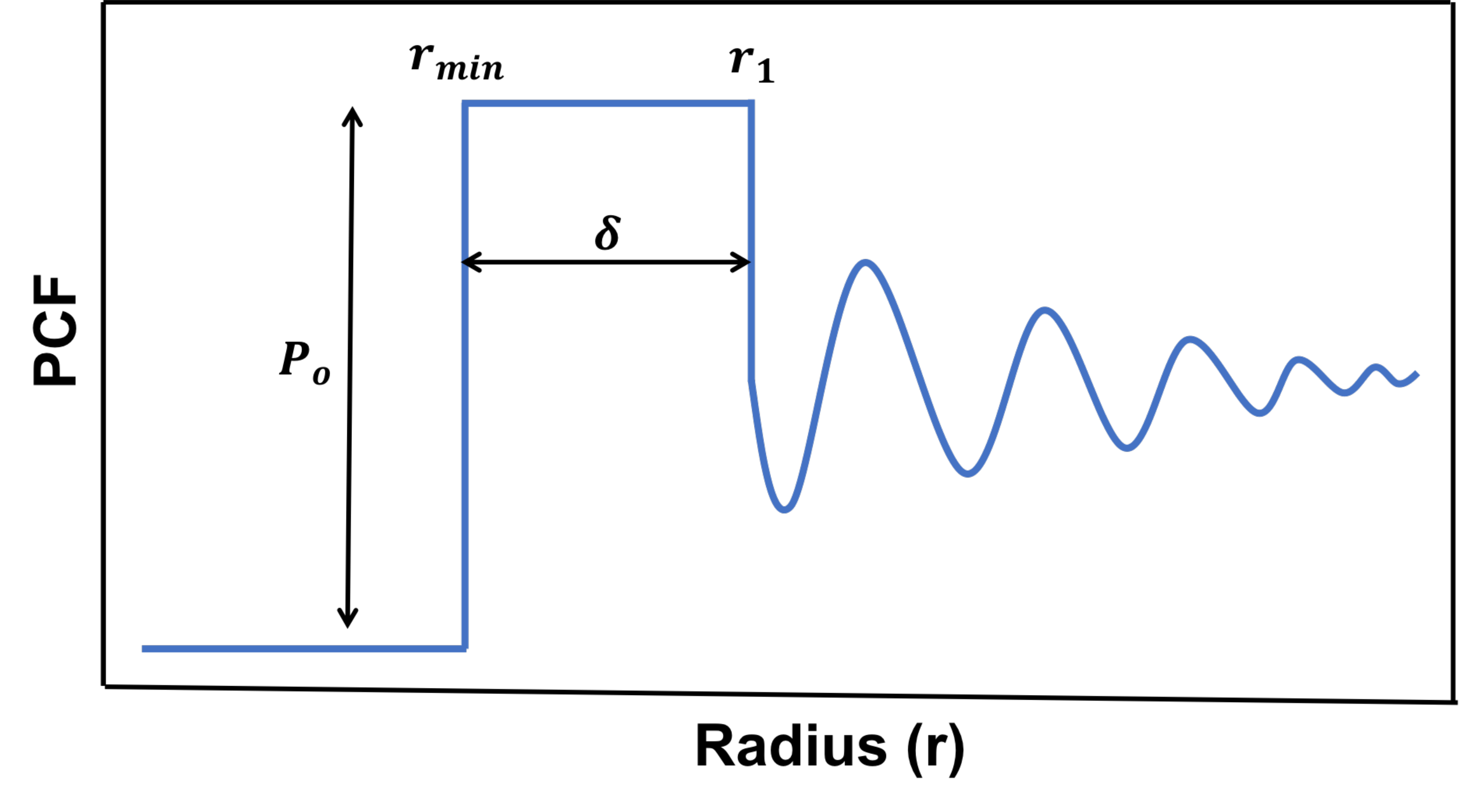}}
	\caption{Pair correlation functions of coverage-based sample designs. Each design leads to a different coverage size $r_{min}$.}
	\label{fig:pcf_coverage}
	\vspace{-0.1in}
\end{figure*}

While Poisson disk sampling was preferred in computer graphics applications for turning regular aliasing patterns into featureless noise, it is not directly suitable for conventional predictive modeling problems. Consequently, in~\cite{kailkhura2018spectral}, the authors studied the impact of coverage on supervised regression problems, and provided empirical evidence that larger coverage in the sample design was critical to improving the expected performance of the models designed using them. Motivated by this observation, they developed the \textit{space-filling spectral design} (SFSD), which is defined as:

\begin{definition}(\textit{Space-Filling Spectral Design})~\cite{kailkhura2018spectral}
\begin{align}
\label{eqn:stair}
& G(r;r_{\text{min}},r_1,P_0)=f(r-r_1)+P_0\left(f(r-r_{\text{min}})-f(r-r_1)\right), \\
\nonumber  & \text{with } f(r-r_{\text{min}}) = \left\{ \begin{array}{rll}
0  & \mbox{if}\ r\leq r_{\text{min}} \\
1  & \mbox{if}\ r>r_{\text{min}}
\end{array} \right\}, \\
\nonumber & \text{where } r_{\text{min}}\le r_1 \text{ and } P_0 \geq 1. 
\end{align}This represents a parameterized stair function (Figure \ref{fig:pcf_coverage}(b)) that introduces a peak in the quest of increasing the coverage.
\end{definition}

\section{Proposed Sample Design Methodology}


In order to enable the effective use of coverage-based designs in ML applications, we need to confront the following challenges: (i) most existing methods are designed specifically for 2-d, and a trivial extension of such constructions provide poor coverage, even in $d>3$; (ii) current sample synthesis algorithms based on PCF matching require extensive manual tuning, and perform poorly as the dimension increases -- in many cases, the synthesis quality is no better than random sampling; and (iii) the superior performance of coverage-based designs has been established mostly on graphics tasks, such as image/volume rendering, and similar gains are yet to be achieved in ML applications. 

In this section, we first propose a new parameterized PCF construction for coverage-based designs, which achieves larger coverage compared to existing approaches. Next, we develop a practical strategy for finding an ``optimal'' PCF configuration. Finally, in the next section, we present an effective sample synthesis algorithm for coverage-based sampling, that consistently leads to high-quality samples across different sample sizes and dimensionality.


\subsection{A New Parameterized Family}
Following notations in the previous section, our parameterized PCF construction for coverage-based sampling can be expressed as follows:
\begin{align}
  \label{eq:proposed}
  & \nonumber G(r;r_{min},r_1,P_0,A,B,C,D)=P_0\left(f(r-r_{min})-f(r-r_1)\right)\\
 & +\left(1+\frac{A}{r}\exp(-Br)\sin(2 \pi Cr+D)\right)*f(r-r_1), \\
  \nonumber  & \text{where } f(r-r_{min}) = \left\{ \begin{array}{rll}
      0  & \mbox{if}\ r\leq r_{min} \\
      1  & \mbox{if}\ r>r_{min}
    \end{array} \right\}, \\ & \nonumber \phantom{where } r_{min}\le r_1 \text{ and } P_0 \geq 1. 
\end{align}The intuition behind this construction is to enable trade-off between randomness and uniformity/coverage properties of a sample design. This construction (see Figure~\ref{fig:pcf_coverage}(c)) has three crucial properties:
\begin{enumerate}
\item The PCF is zero from $0 \leq r \leq r_{min}$, corresponding to the coverage size similar to other designs;
\item The PCF has a peak from $r_{min} < r \leq r_1$ and damped oscillations from $r>r_1$ characterizing randomness;
\item The peak height $P_0$, width $\delta = r_1-r_{min}$, and oscillations can be adjusted to control the randomness property of a design, which in turn can maximize the coverage $r_{min}$.
\end{enumerate}The radially-averaged power spectral density of the PCF in \eqref{eq:proposed} can be obtained using the relation in \eqref{eq:link}. As we will show later, this connection is central for designing and optimizing the proposed design in a computationally efficient manner.


\begin{figure}[t]
	\centering
	
	\subfigure[Fixed $N=1000$ and varying $d$]{\includegraphics[width=.48\linewidth]{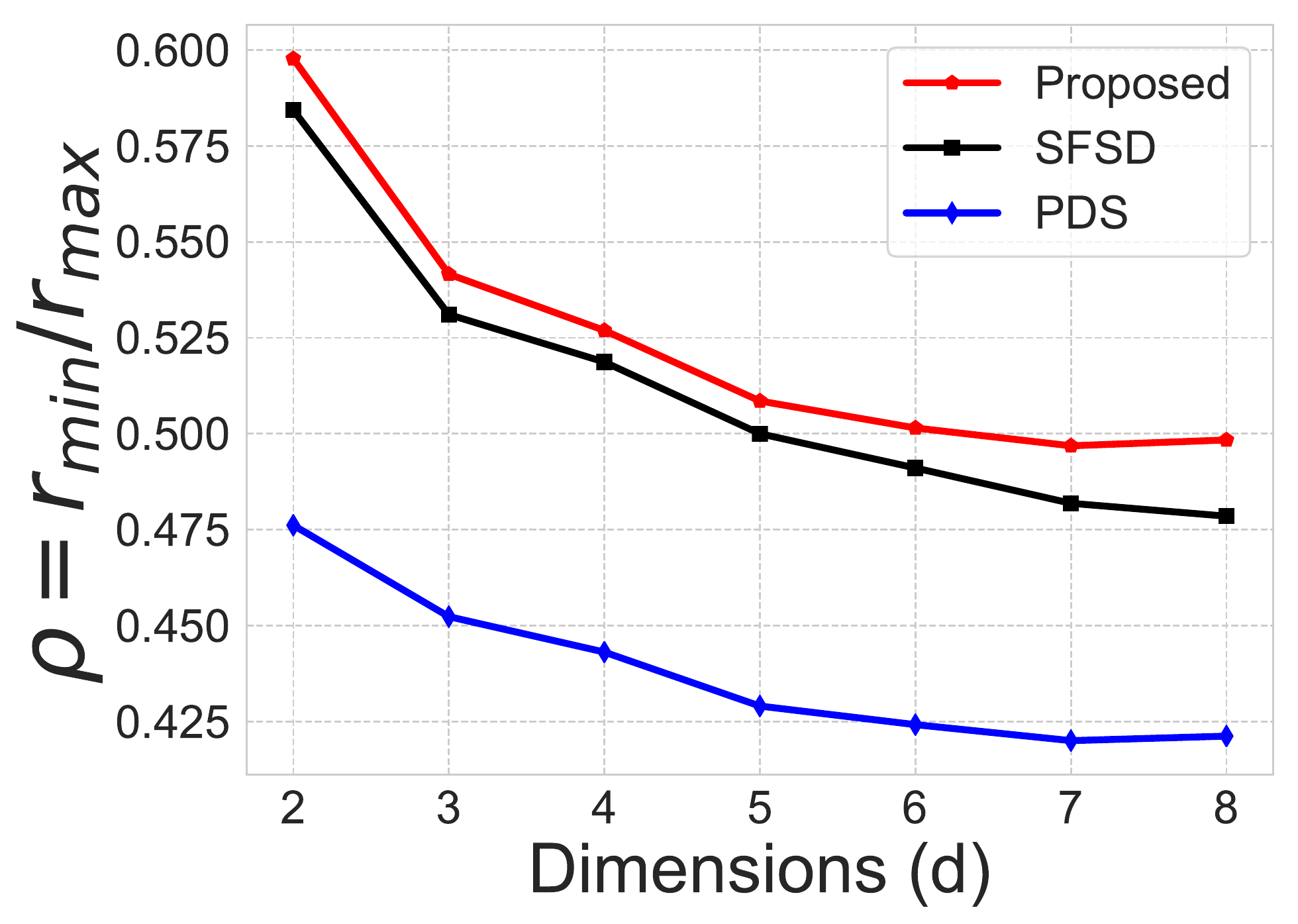} \label{fig:rho_fixN}
	}
\subfigure[Fixed $d=5$ and varying $N$]{\includegraphics[width=.48\linewidth]{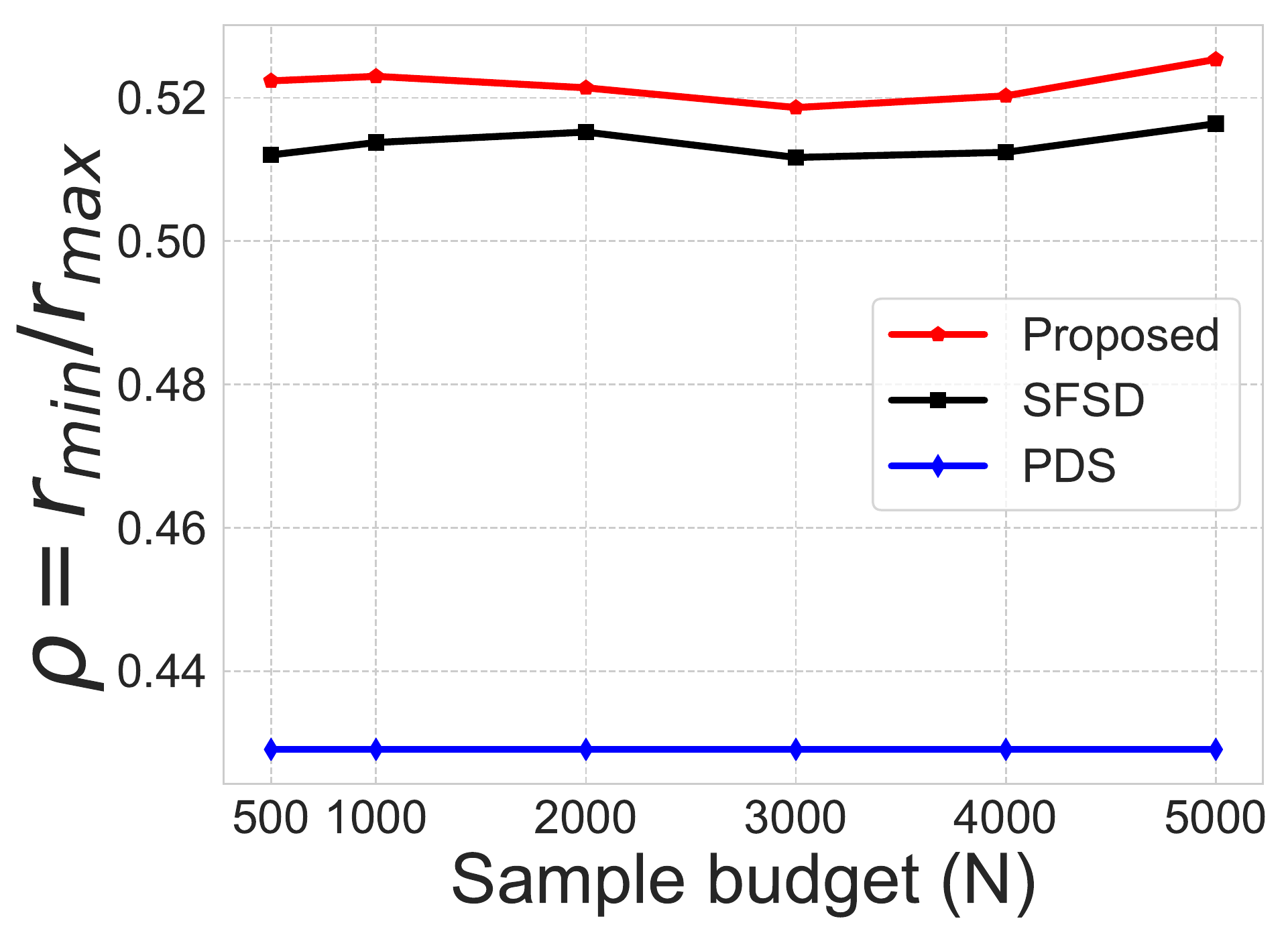} \label{fig:rho_fixd}
} 
	\caption{Maximum achievable \textit{relative radius} using different coverage-based designs. The proposed design consistently outperforms PDS and SFSD approaches.} 
	\label{fig:rho}
\end{figure}

\subsection{Quantifying Coverage Gain}
Next, we evaluate the coverage gain in our proposed design with respect to other coverage-based approaches from the computer graphics and surrogate modeling literature. 
For this analysis, we varied the parameter $P_0$ in the range $[1.00, 2.5]$ and performed a brute-force search on the parameter $r_1\in[r_{\rm min}, 2r_{\rm min}], A\in[0.1,0.9], B\in[2,6],C\in[50,600],D\in[-\pi,\pi]$, such that $r_{\rm min}$ is maximized, while also ensuring that the realizability conditions from Definition \ref{real} are met.



We compare the coverage characteristics of the proposed approach to existing coverage-based designs, namely PDS and SFSD. More specifically, we compare different coverage-based designs using the \textit{relative radius} \cite{gamito2009accurate} $\rho = r_{\rm min}/r_{\rm max}$, where
$r_{\rm max}$ is the maximum possible radius for $N$ samples in $d$ dimensions. For a given $N$ and $d$, $r_{\rm max}$ can be computed as
$$
r_{\rm max} = \sqrt[d]{\gamma_d \frac{V~\Gamma\left(\frac{d}{2}+1\right)}{\pi^{\frac{d}{2}}N}},
$$where the maximum packing density $\gamma_d$ for $d=\{2,\cdots,8\}$ can be found in \cite{online_packingdensity}. 

In Figure~\ref{fig:rho_fixN}, we first fixed the the sample size at $N=1000$. Subsequently, we measured the maximum achievable relative radius using different coverage-based designs of size $N$ in dimensions $d=\{2,\cdots,8\}$ respectively. The first striking observation is that by incorporating controlled randomness, both SFSD and the proposed design produce significantly larger relative radius when compared to the conventional PDS. Further, in all cases, the proposed approach provides improved coverage over SFSD and as we will demonstrate in our results, this seemingly marginal improvement leads to significant performance gains in practice. Another interesting observation is that as the dimensionality increases, the relative radius decreases rapidly and all coverage-based design behave similarly. In other words, due to the curse of dimensionality, when the search space is comprised of tens of dimensions, the proposed approach will become similar to PDS (unless the sample size grows exponentially), while still being superior to discrepancy-based designs. 

In practice, with applications such as hyper-parameter optimization, unless the intrinsic dimension of the optimization surface is low, exploratory sampling will be ineffective (even with million samples) as the volume of spaces grows exponentially with dimension. In the literature, it has been observed that the intrinsic dimension of search spaces is often between $3-6$ over different datasets~\cite{bergstra2012random}. Similarly, in Figure~\ref{fig:rho_fixd}, we show the relative radius $\rho$ at different sample sizes for a given dimension $d=5$. As it can be seen, for all coverage-based designs, the best achievable relative radius is nearly a constant at all sample sizes. Furthermore, the proposed design consistently produces larger coverage compared to other designs in all cases.


\begin{algorithm}[t]
	\caption{Automatic selection of PDS parameters}\label{alg:auto_par}
	\begin{algorithmic}[1]
		\State \textbf{Input:} Number of samples $N$, dimension $d$,  parameter $P_0$, step size $\lambda$, $V = 1$.
		\State $\bar{r}_{min}  = \sqrt[d]{{V\Gamma\left({d}/{2}+1\right)}/(\pi^{\frac{d}{2}}N)}$ \Comment{Conventional PDS}
		\State \textbf{Initialize:} $r_{min} = \bar{r}_{min}, r_1 = 2 r_{min}$
		\State $G(r) \gets G(r;\;r_{min},r_1,P_0)$ \Comment{Initialize PCF}
		\State $P(k) \gets 1+\dfrac{N}{V} (2\pi)^{\frac{d}{2}}k^{1-\frac{d}{2}}H_{\frac{d}{2}-1}\left[r^{\frac{d}{2}-1}G(r)-1\right]$ \Comment{from \eqref{eq:link}}
		\State $k^* \gets \underset{k}{\rm \arg\min}\;P(k)$
		\State \textbf{While} $({P}(k^*) \geq 0)$   \Comment{Constraint for realizable PCF}
		\State \hspace{0.5cm} Update $r_{min}$ \\
		 \hspace{1cm} $r_{min} \gets r_{min} + \lambda \bigg( r_{min} \dfrac{\partial}{\partial r_{min}} P(k^*)+P(k^*) \bigg) $ 
		\State \hspace{0.5cm} Update $r_1$ \\
		\hspace{1cm} $r_1 \gets r_1 - \lambda \bigg( r_{min}\dfrac{\partial}{\partial r_1} {P}(k^*) \bigg)$
		\State \textbf{Return} $r_{min},r_1$ \Comment{Optimal PCF settings}
	\end{algorithmic}
\end{algorithm}

\subsection{A Practical Strategy for Parameter Selection}
A typical approach to find parameters that achieve the largest coverage gain is a brute-force search~\cite{kailkhura2018spectral}. However, the search space of realizable parameters is complex -- non-monotonic, coupled with the need to satisfy realizability conditions. To overcome this challenge, we develop an efficient gradient based parameter selection strategy for optimal PCF construction. 
Specifically, we are interested in solving the following parameter search problem\footnote{{{We found that the maximum achievable coverage, $r_{min}$, for a given $d$ and $N$, depends primarily on the choice of $r_1$ and $P_0$, while the choices for other parameters $A, B, C, D$ are not particularly sensitive. Thus, we only optimize over $r_1$ and $P_0$ in this paper.
}}}:
\begin{align}\label{eqn:opt_prob}
{\rm maximize}~:&~ r_{\rm min} \nonumber\\
{\rm subject~to~:}&~P(k)\ge 0, \forall k \nonumber\\
&~ r_1 > r_{\rm min}
\end{align}
Since the goal is to achieve maximal coverage, we maximize $r_{min}$ such that the resulting PCF is realizable, which is verified by ensuring that the power spectral density $P(k)\geq 0 , \forall k$. 
In our experiments, we found that the lagrangian relaxation of \eqref{eqn:opt_prob} is hard to optimize. Instead, we maximize an alternative objective function $r_{min}\times P(k^*)$ where $k^*=\arg\min P(k)$, ({i.e.,} consider only the  minimum value of $P(k)$) is found to work better. In Algorithm \ref{alg:auto_par}, we present our approach to solve this modified optimization problem. 


\section{Proposed Synthesis Algorithm}
\begin{algorithm}[!t]
	\caption{Sample Design using GD-ALR Algorithm}\label{alg:pcf}
	\begin{algorithmic}[1]
		\State \textbf{Input:} Number of samples $N$, dimension $d$, target PCF $\hat{G}^*(r_j)$, learning rate $\lambda$
		\State $\mathbf{X}\gets \text{Random} (N,d)$ \Comment{Initial random sample design}
		\State $G\gets \text{PCF}(\mathbf{X})$ \Comment{Calculate initial PCF using Eq.~\eqref{eqn:pcf_estimator}} 
		\For{$t=1$ to $T$}\Comment{Total $T$ gradient descent iterations}
		\For{$i=1$ to $N$} \Comment{Update each sample at a time}
		\State $\Delta_i^p\gets \dfrac{\partial}{\partial x_i^p} \sum\limits_{j=1}^{M}\left({G^t(r_j)-G^*(r_j)}\right)^2\; \text{for}\; p\in\{1,\cdots,d\}$ \Comment{Calculate gradients}
		\State $\lambda \gets 0.1 e^{-0.1 \sqrt{t}}$ \Comment{ Adapting learning rate}
		\State $\mathbf{x}_i^p(t+1)\gets \mathbf{x}_i^p(t)-\lambda \dfrac{\Delta_i^p}{|\Delta_i^p|}$\Comment{Update the samples position} 
		\State $G^t\gets \text{PCF}(\mathbf{X})$\Comment{Update the PCF}
		\EndFor
		\EndFor
		\State \textbf{return} $\mathbf{X}$\Comment{Optimized Samples}
	\end{algorithmic}
\end{algorithm}

We develop an approach that iteratively transforms an initial random sample design such that its PCF matches the PCF of the optimal coverage-based design. More specifically, we consider a non-linear least squares formulation similar to~\cite{Oztireli:2012,kailkhura2018spectral}. Despite being computationally efficient, due to the high non-convexity of the PCF matching problem, conventional gradient descent based approaches perform poorly as the dimension increases. In fact, due to the small effective $r_{min}$, the synthesis quality is no better than random sampling. Here, we adopt a different approach to alleviate this limitation, and make PCF matching-based synthesis a viable solution.

Denoting the desired PCF for an optimal design by $G^*(r)$, we discretize the radius $r$ into $M$ points $\{r_j\}_{j=1}^{M}$, and minimize the sum of the weighted squares of errors between the target PCF $G^*(r_j)$ and the curve-fit function (explained next) $G(r_j)$. Consequently, sample synthesis is posed as the following non-linear least squares problem:
\begin{equation}
	\min~ \sum\limits_{j=1}^{M}\left(G(r_j)-G^*(r_j)\right)^2.
\end{equation}

%

\subsection{PCF Matching Algorithm}
Intuitively, the proposed PCF matching algorithm is comprised of two phases: achieving coverage and matching oscillations. In the first phase, the initial design of uniform random samples is optimized to achieve coverage by shifting the positions of the $N$ samples, such that no two samples are closer than $r_{min}$. In the second phase, samples are optimized to match oscillations in the target PCF. Before presenting the proposed algorithm, we first describe the PCF estimator employed in our optimization.

\noindent \textbf{PCF Estimator:} To estimate the PCF of point samples, we employ a kernel density estimator~\cite{Oztireli:2012}, defined as 

\begin{equation}
\label{eqn:pcf_estimator}
 \hat{G}(r)=\frac{V_W}{\gamma_W}\frac{V_W}{N}\frac{1}{S_E
	(N-1)}\sum\limits_{i=1}^{N}\underset{i\neq
	j}{\sum\limits_{j=1}^{N}}k\left(r-|\mathbf{x}_i-\mathbf{x}_j|\right)
\end{equation}
where $k(.)$ denotes the Gaussian kernel function, $k(z)=({1}/{\sqrt{\pi}\sigma})\exp(-z^2/{2\sigma^2}).$ In this expression, $V_W$ indicates the volume of the search space and $S_E$ denotes the area of hyper-sphere. Finally, $\gamma_W$ is an isotropic set covariance function
which can be approximated as   $\gamma_W=V_W-(S_W/\pi) r$,
where $S_W$ denotes the surface area of the sampling region.
The term $\frac{V_W}{\gamma_W}$ accounts for edge correction for the unboundedness of the estimator. 

\begin{figure}[t]
 	\centering
 	\subfigure[]{\includegraphics[width=.48\linewidth]{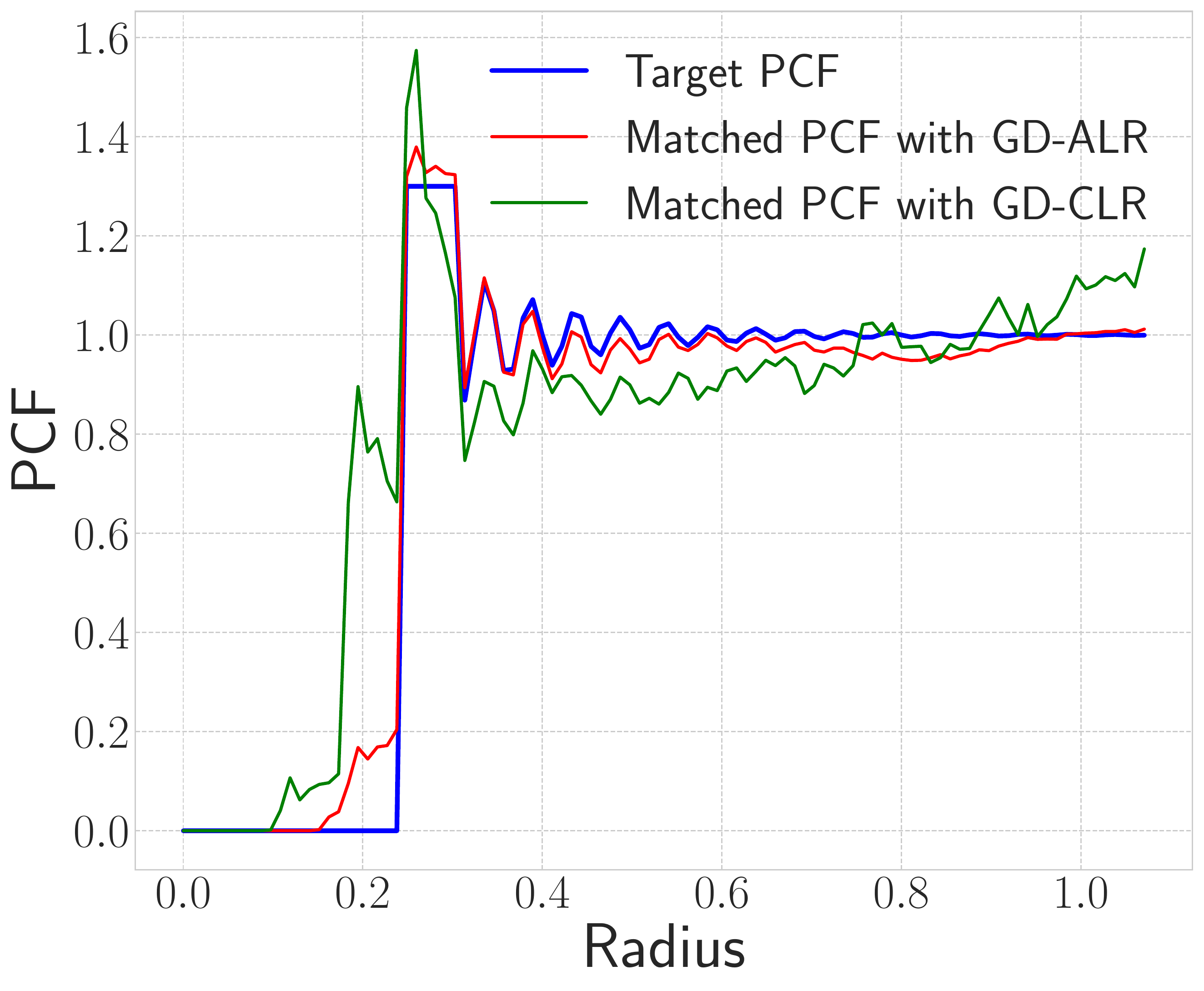} \label{fig:proposed_samp}
 	}
 	\subfigure[]{\includegraphics[width=.48\linewidth]{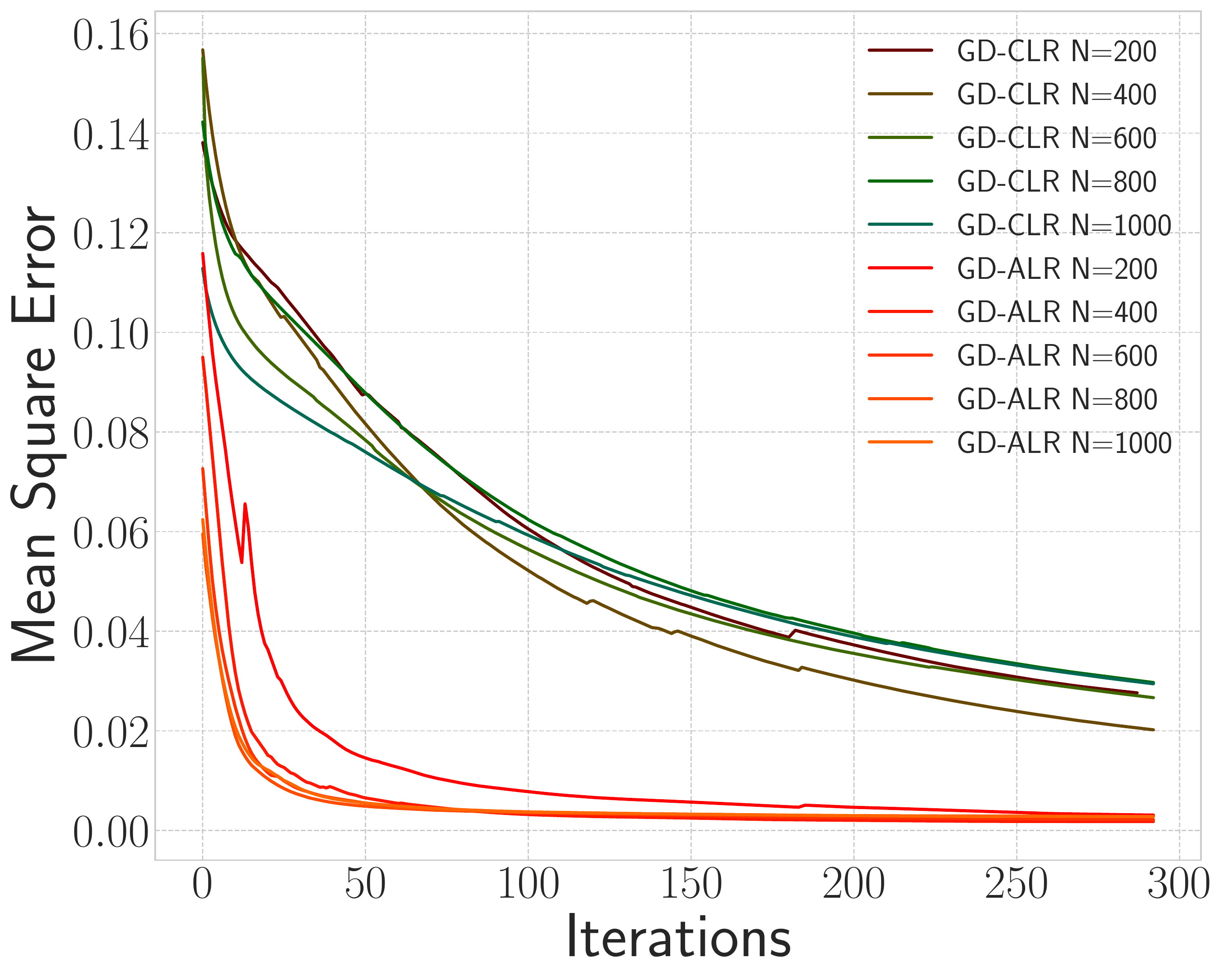} \label{fig:adamsgd}
 	}
 	\caption{Coverage-based Sample Synthesis. (a) PCF matching performance of GD-ALR versus GD-CLR, while obtaining maximal coverage ($r_{min}$) for $N=200, d=4, P_0 = 1.3$. (b) Mean square error of PCF matching obtained using GD-ALR and GD-CLR, across different gradient descent iterations, for a fixed dimension $d=4$. } 
 	\label{fig:syn_results}
 \end{figure}

\noindent \textbf{Algorithm:} Given the PCF estimate, the matching problem can be solved using gradient descent. However, due to the highly non-convex nature of this problem, gradient decent with constant learning rate (GD-CLR)~\cite{kailkhura2018spectral} perform very poorly. Instead, we propose to employ gradient descent with adaptive learning rate, GD-ALR (Algorithm~\ref{alg:pcf}), with the learning rate update rule: $\lambda = 0.1 e^{-0.1 \sqrt{t}},$ for iteration $t$. More importantly, we find that, in order to achieve the maximal coverage, it is important to first optimize for coverage (updates with larger values of $\lambda$) and, then for oscillations (updates with smaller value of $\lambda$), instead of joint optimization as done in existing approaches \cite{kailkhura2018spectral,Heck:2013}. This behavior is illustrated in Fig.~\ref{fig:proposed_samp}. Separately optimizing for coverage/oscillations using an adaptive learning rate profile, solves a major bottleneck in synthesizing coverage-based designs. In particular, we found that many other variants of gradient descent (e.g. Levenberg–Marquardt) failed to achieve the desired performance.
 
From Fig.~\ref{fig:adamsgd}, it can be seen that the proposed GD-ALR demonstrates superior convergence characteristics when compared to GD-CLR.
We also conducted experiments with other optimization approaches, such as, momentum gradient descent optimizer.  We observed that in all settings of $N$ and $d$, GD-ALR outperformed other optimizers with faster convergence and significantly lesser PCF matching error. In summary, the proposed improvements to sample synthesis enables unprecedented capabilities in exploratory sampling. We demonstrate that using experiments in predictive modeling and hyper-parameter optimization.



\section{Empirical Study: Sampling for Predictive Modeling}


\begin{figure*}[!t]
	\centering
	\subfigure[Alpine $N.1$ ($d=3$)]{%
		\includegraphics[width=.3\textwidth]{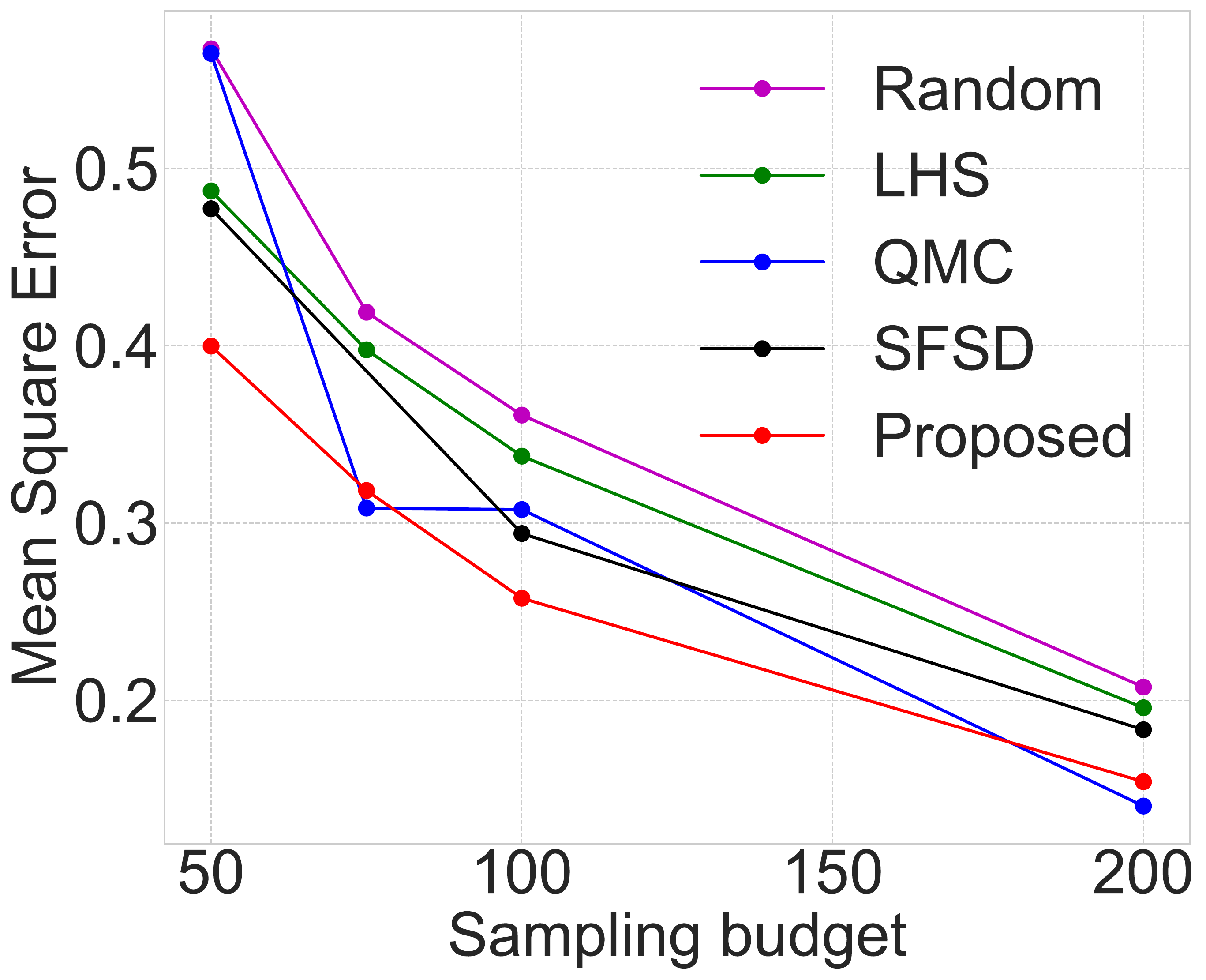} \label{fig:Blind_alpine_3D}
	}
	\subfigure[Alpine $N.1$ ($d=4$)]{%
		\includegraphics[width=.3\textwidth]{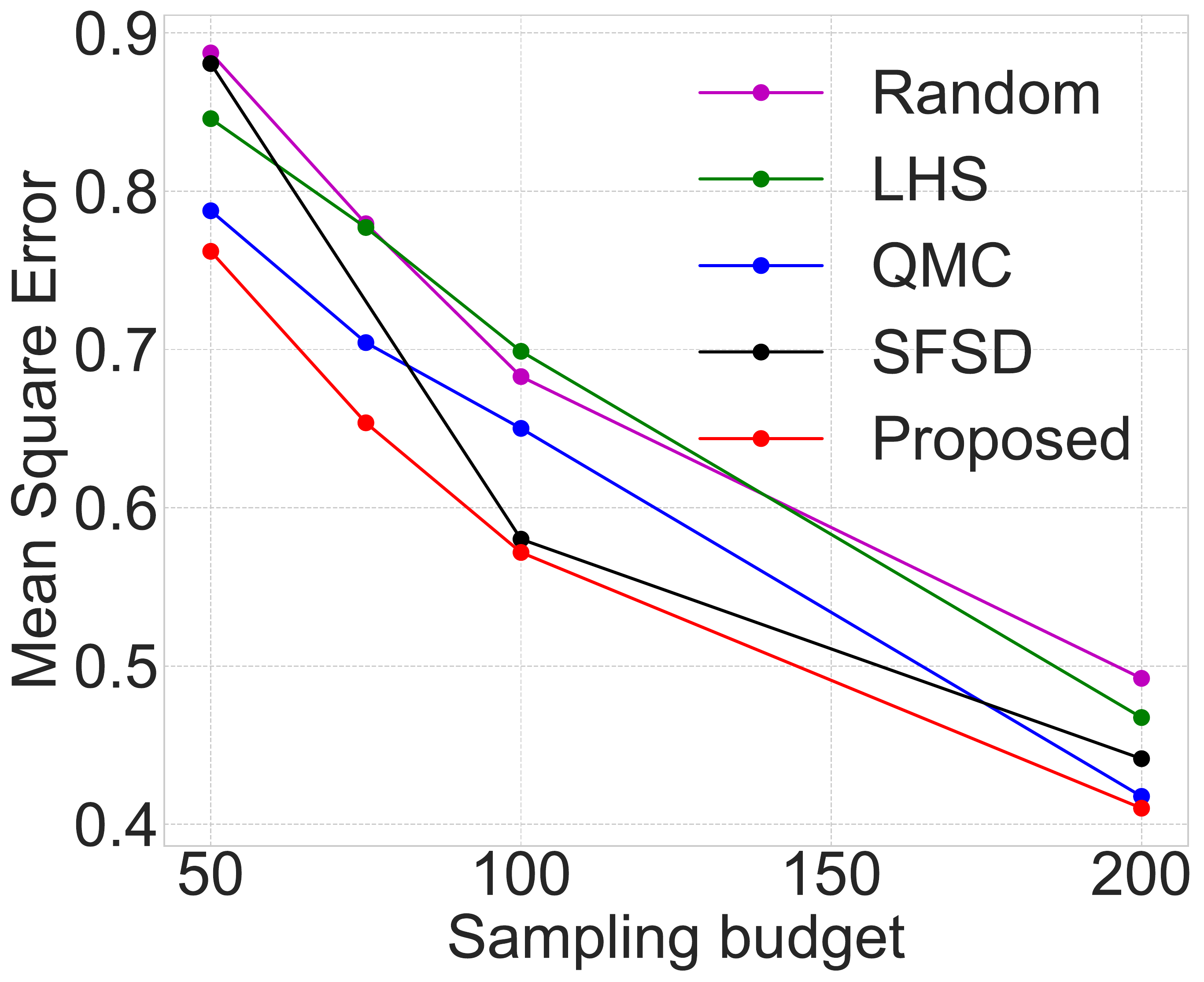} \label{fig:Blind_alpine_4D} 
	} 
	\subfigure[Alpine $N.1$ ($d=5$)]{%
		\includegraphics[width=.3\textwidth]{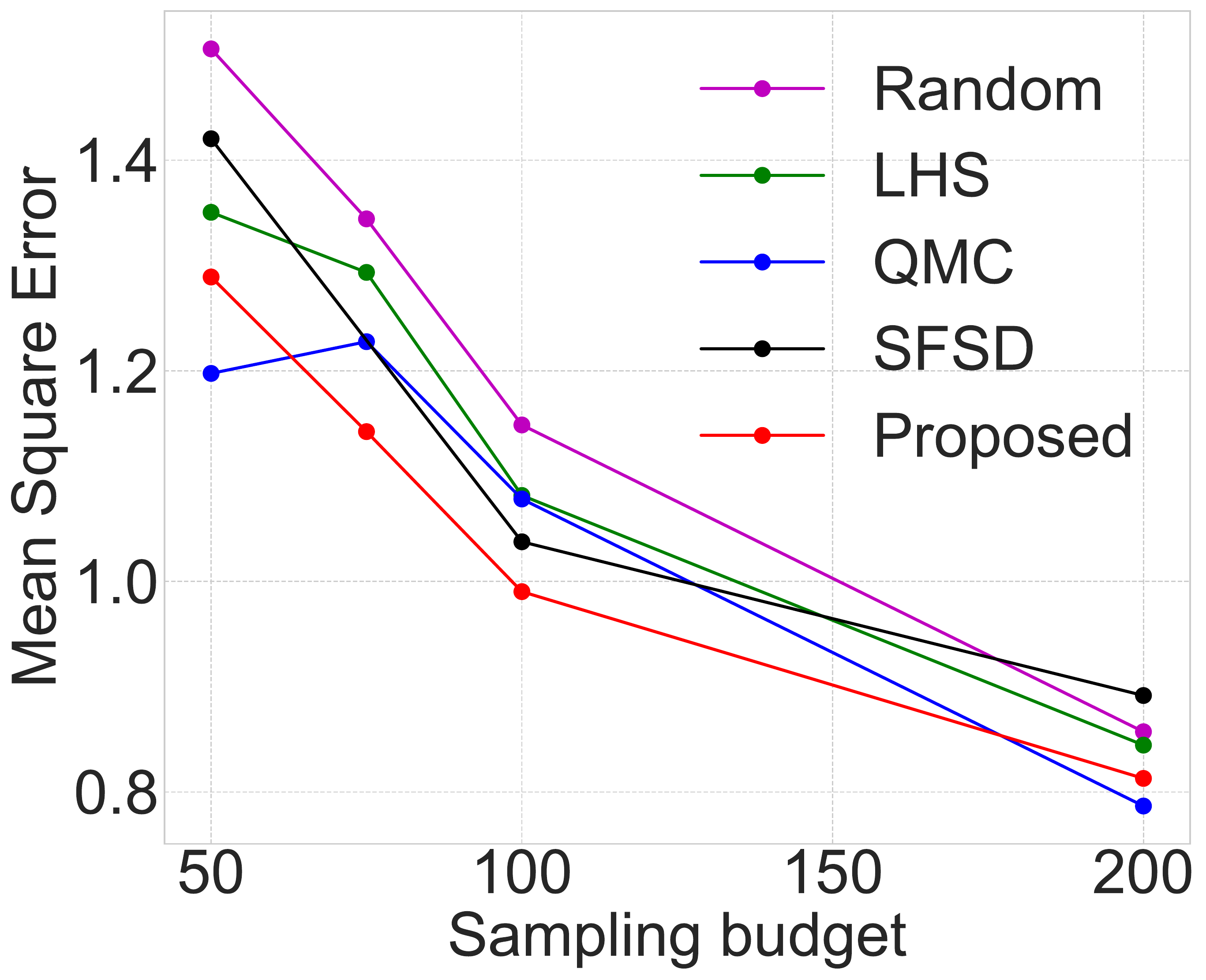} \label{fig:Blind_alpine_5D} 
	}
	\subfigure[Ackley ($d=3$)]{%
		\includegraphics[width=.3\textwidth]{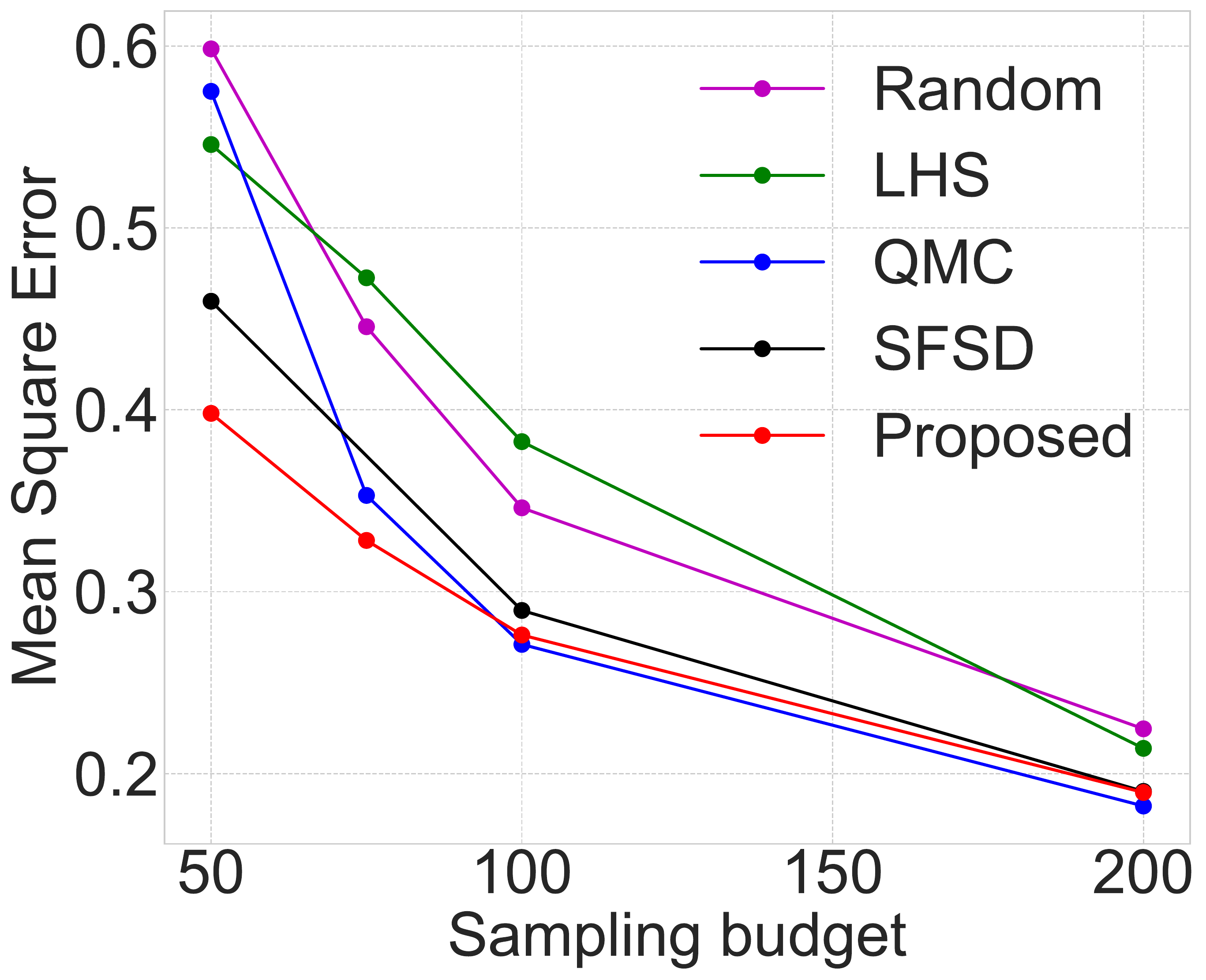} \label{fig:Blind_ackley_3D}
	}
	\qquad
	\subfigure[Ackley ($d=4$)]{%
		\includegraphics[width=.3\textwidth]{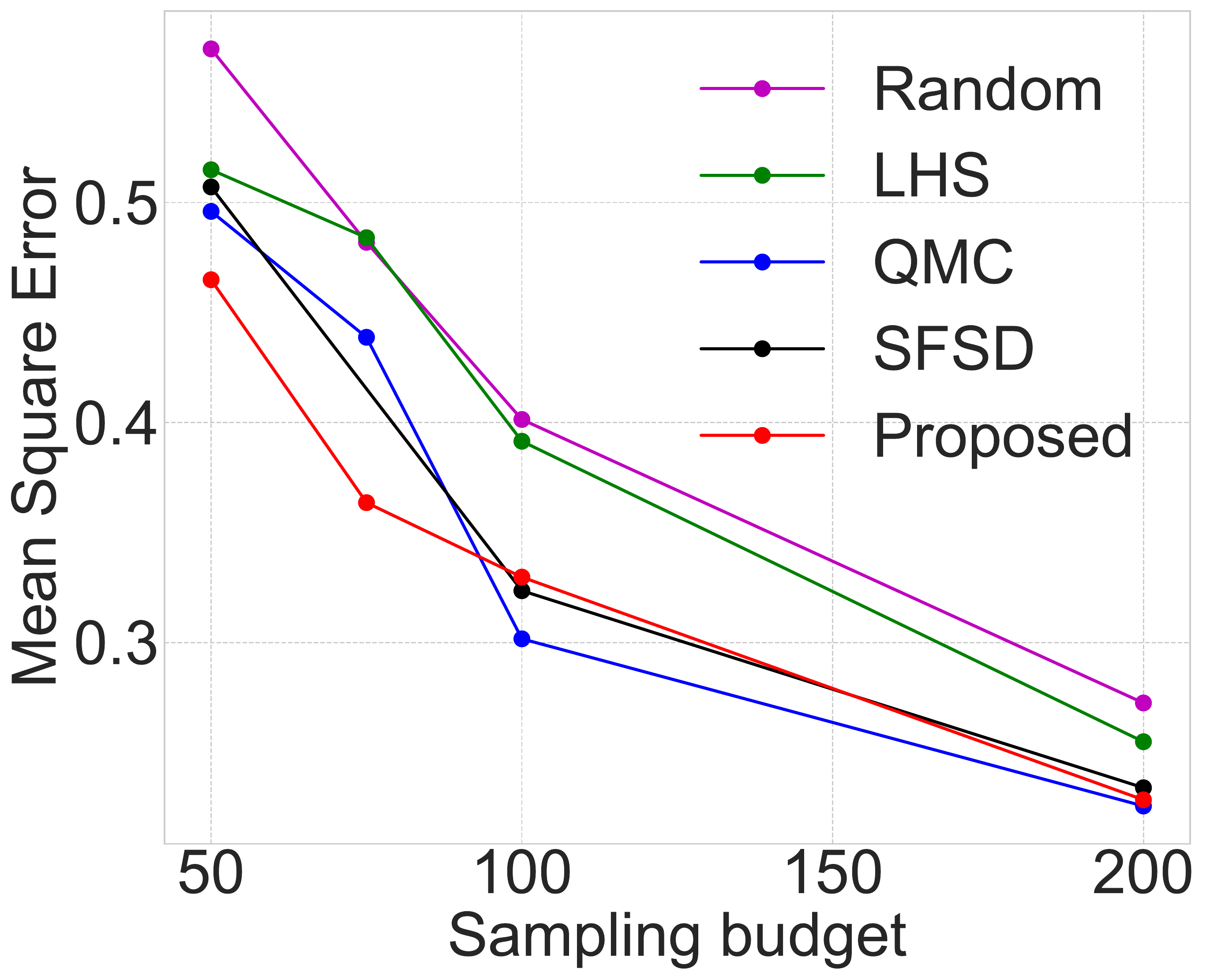} \label{fig:Blind_ackley_4D} 
	} 
	\subfigure[Ackley ($d=5$)]{%
		\includegraphics[width=.3\textwidth]{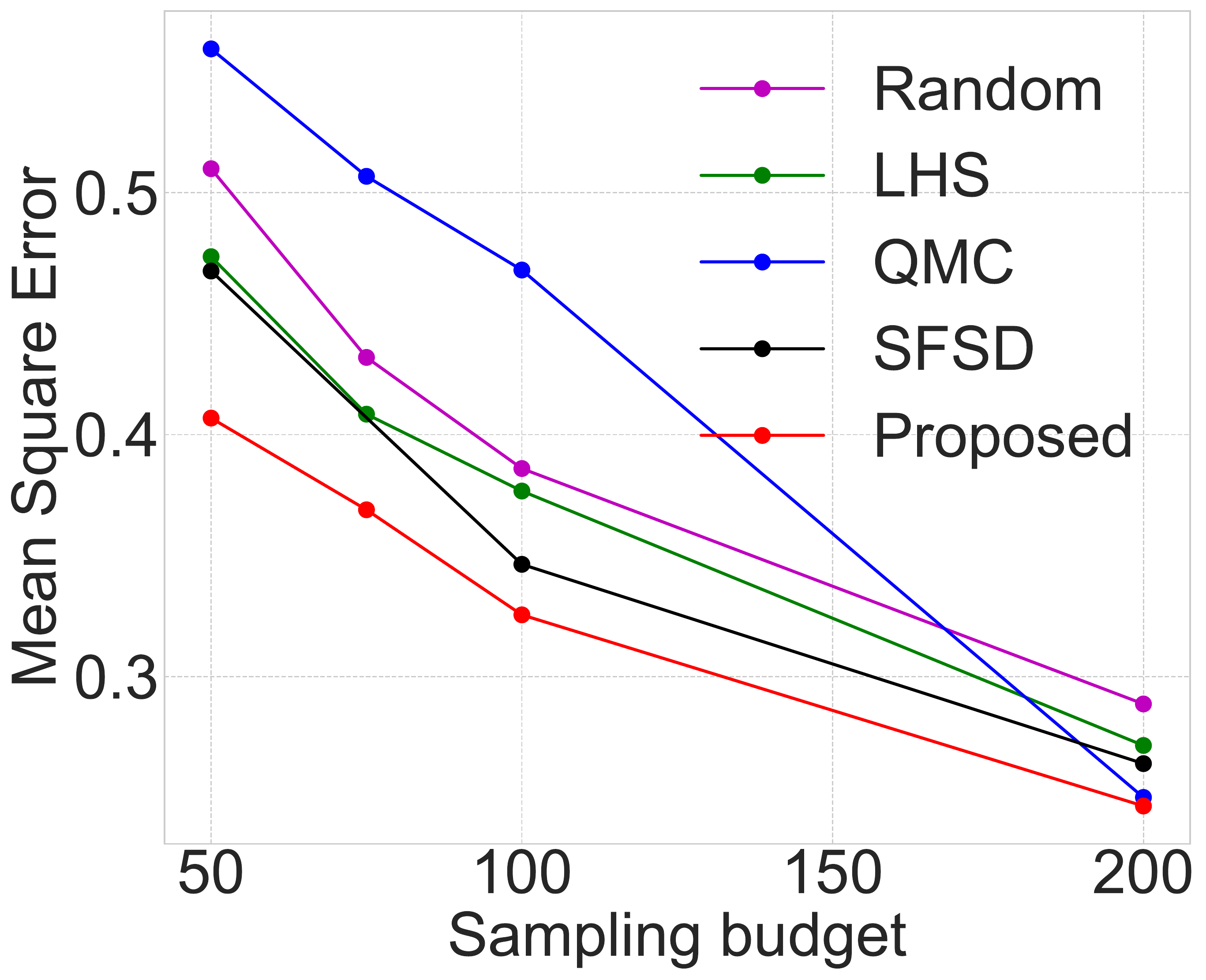} \label{fig:Blind_ackley_5D} 
	}
	\subfigure[Alpine $N.1$ ($d=3$)]{%
		\includegraphics[width=.3\textwidth]{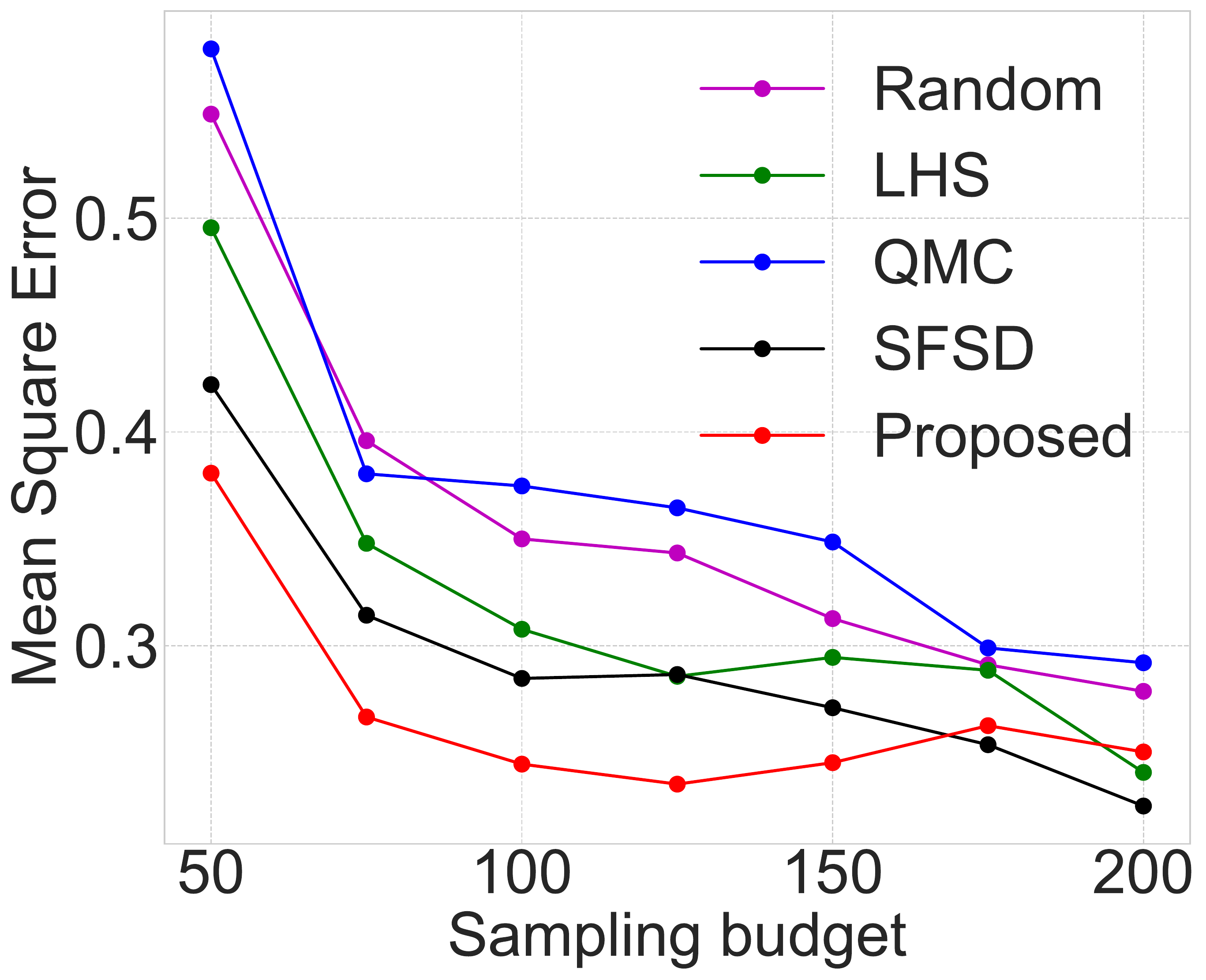} \label{fig:Bayes_alpine_3D}
	}
	\subfigure[Alpine $N.1$ ($d=4$)]{%
		\includegraphics[width=.3\textwidth]{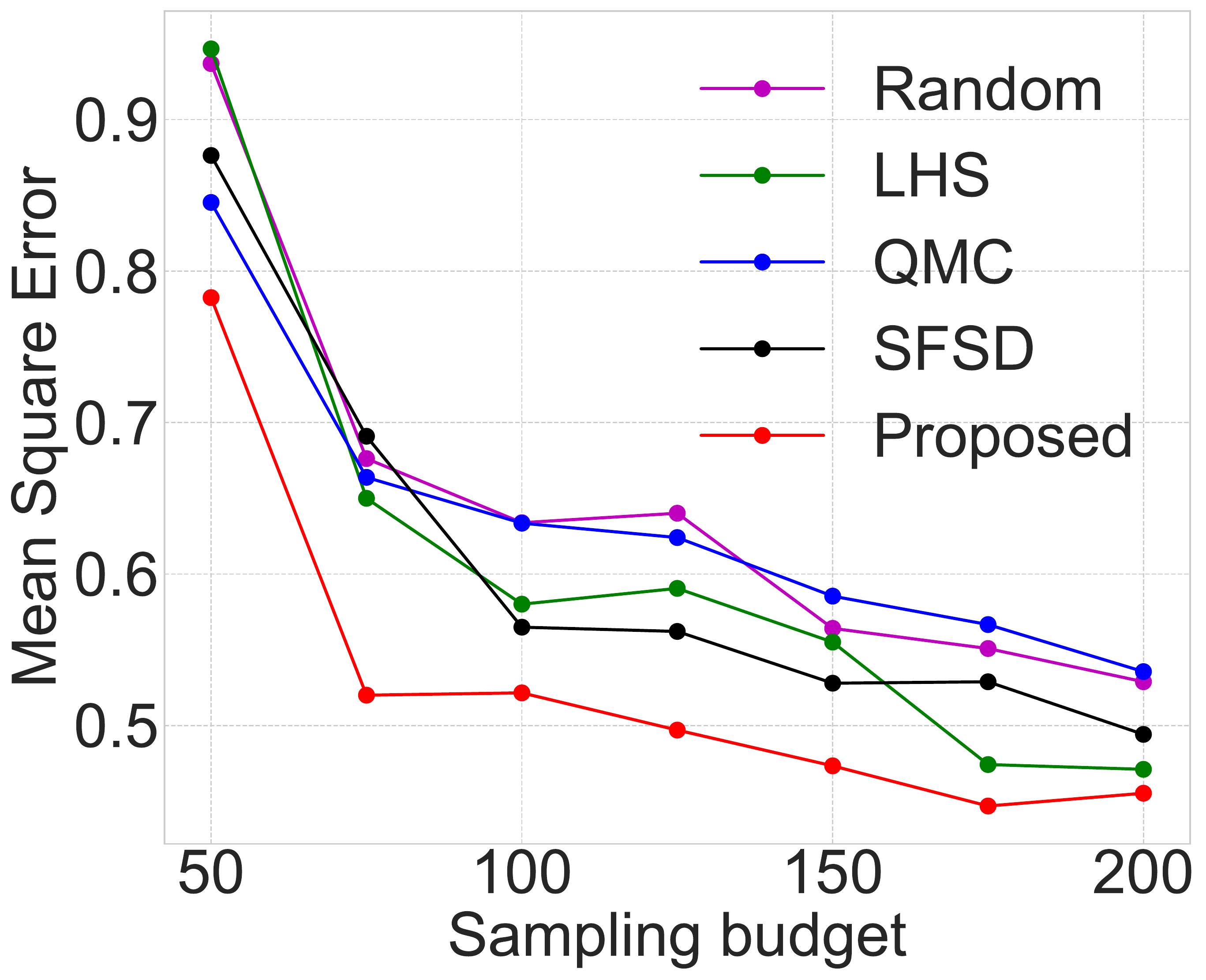} \label{fig:Bayes_alpine_4D} 
	} 
	\qquad
	\subfigure[Alpine $N.1$ ($d=5$)]{%
		\includegraphics[width=.3\textwidth]{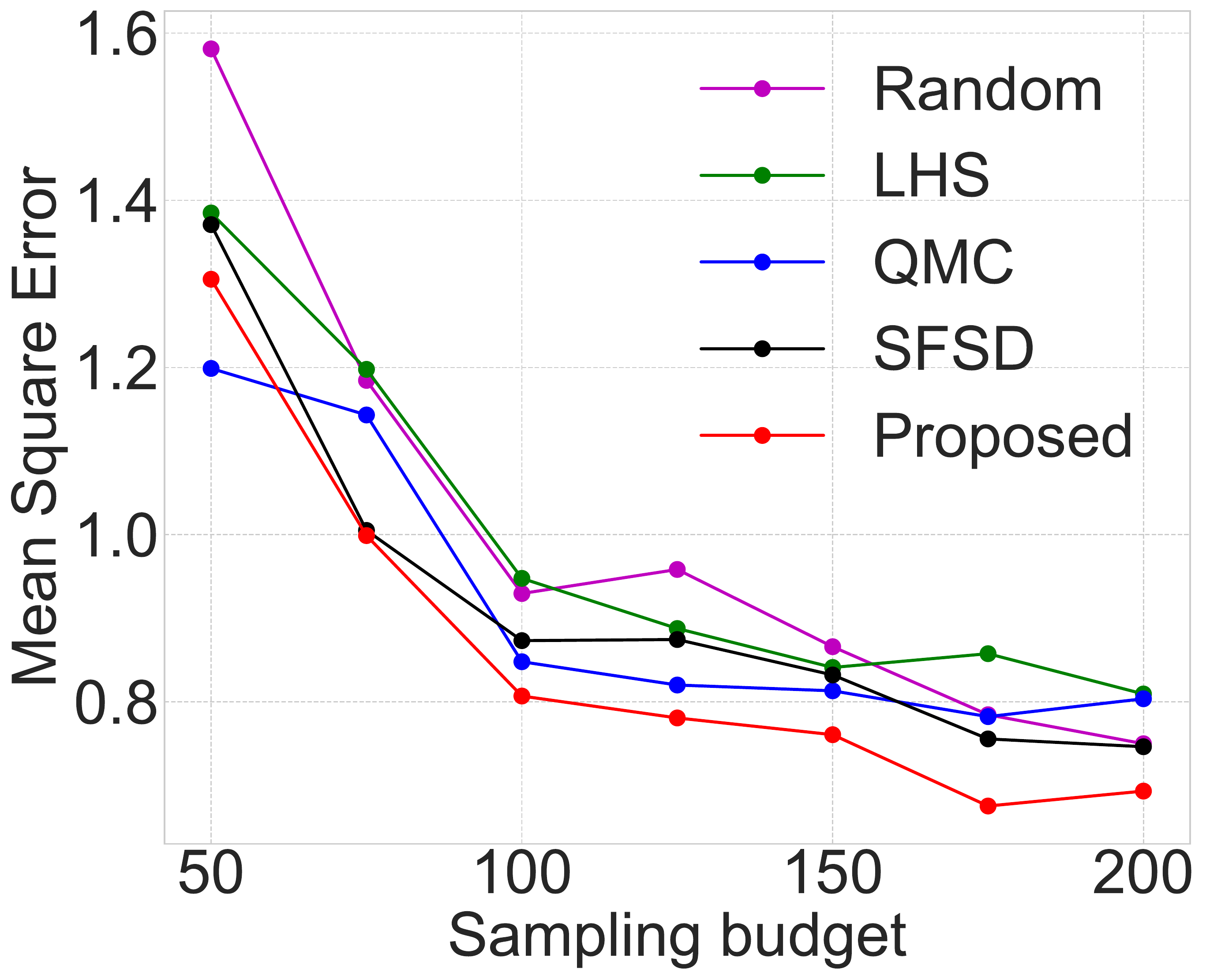} \label{fig:Bayes_alpine_5D} 
	}
	\subfigure[Ackley ($d=3$)]{%
		\includegraphics[width=.3\textwidth]{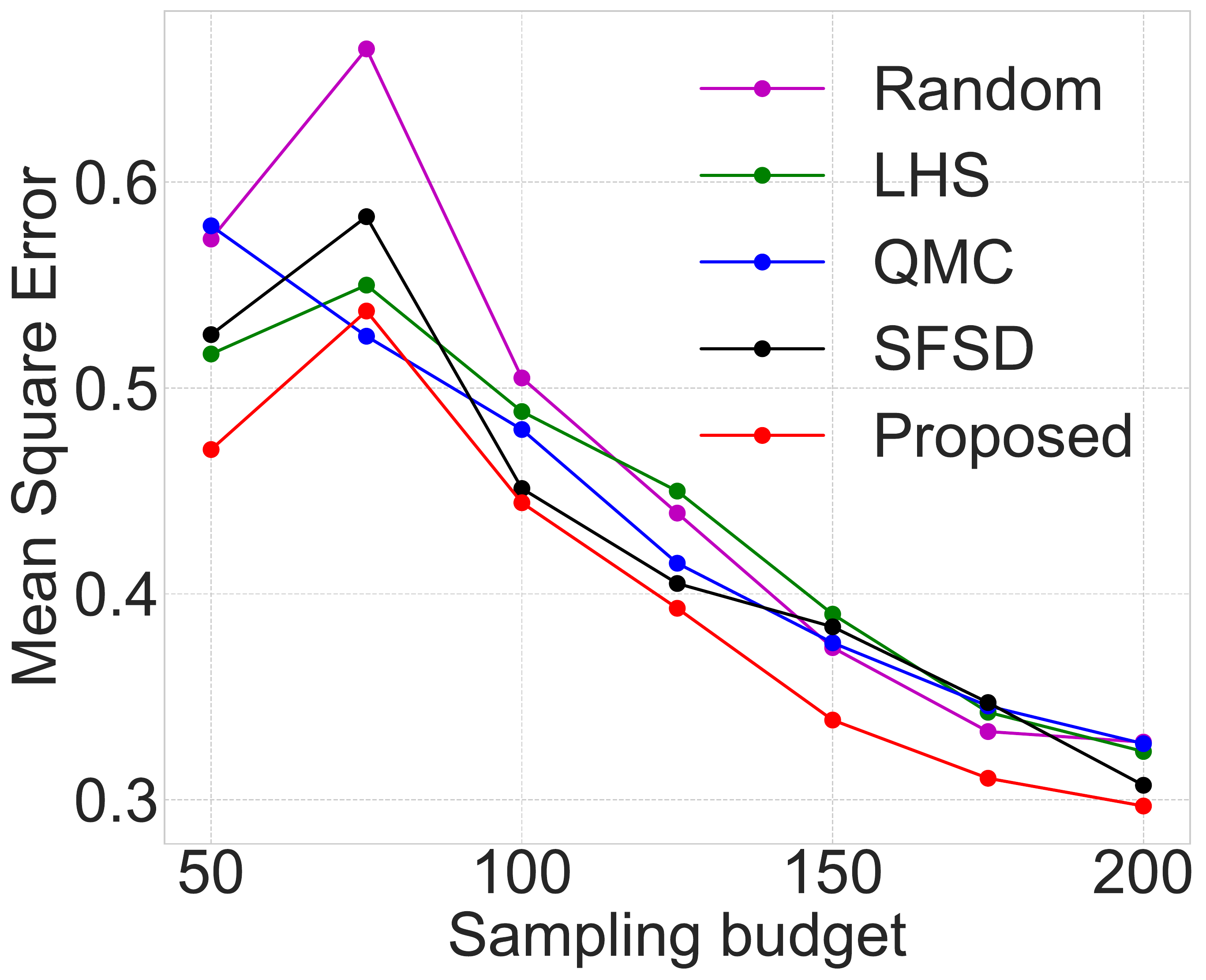} \label{fig:Bayes_ackley_3D}
	}
	\subfigure[Ackley ($d=4$)]{%
		\includegraphics[width=.3\textwidth]{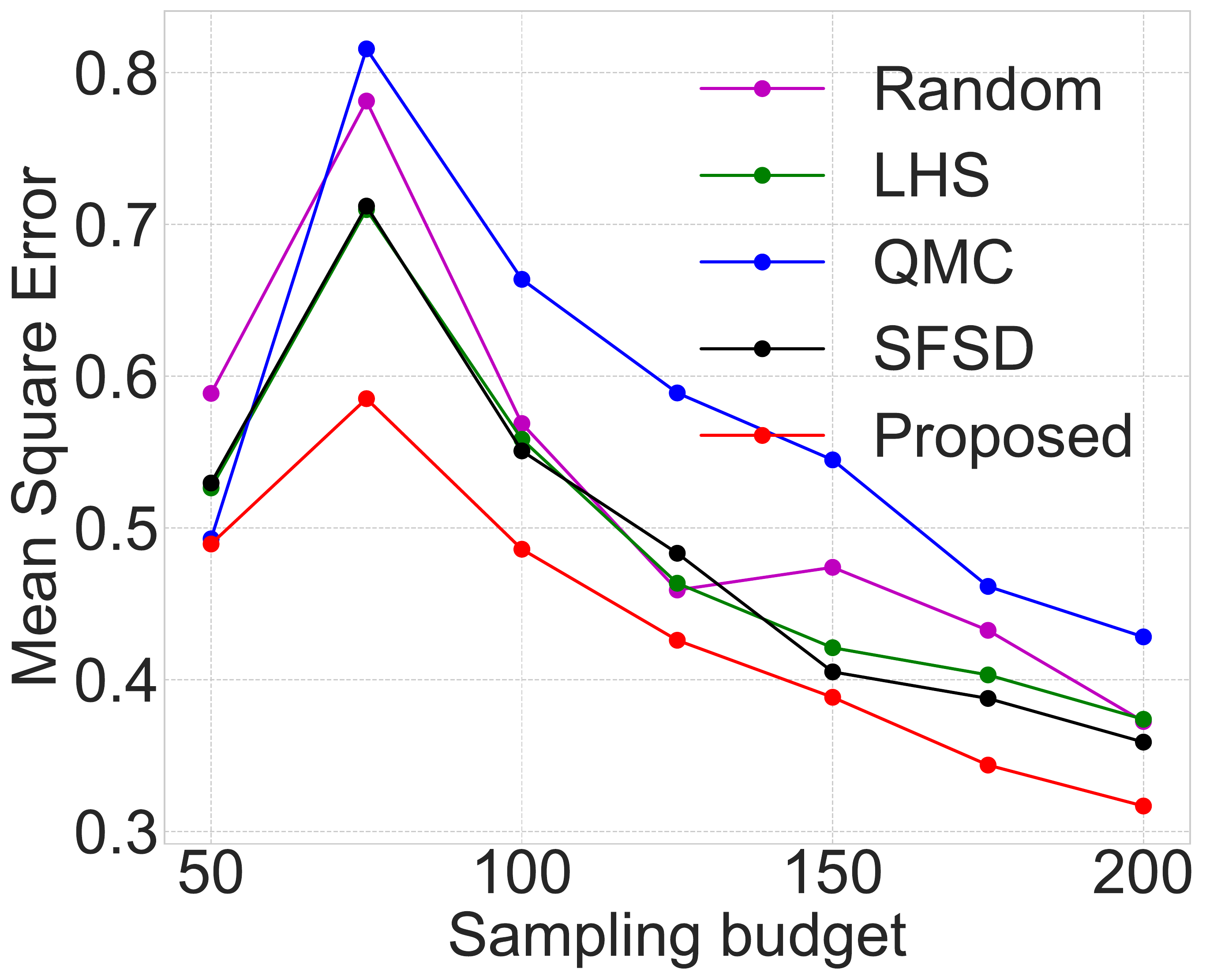} \label{fig:Bayes_ackley_4D} 
	} 
	\subfigure[Ackley ($d=5$)]{%
		\includegraphics[width=.3\textwidth]{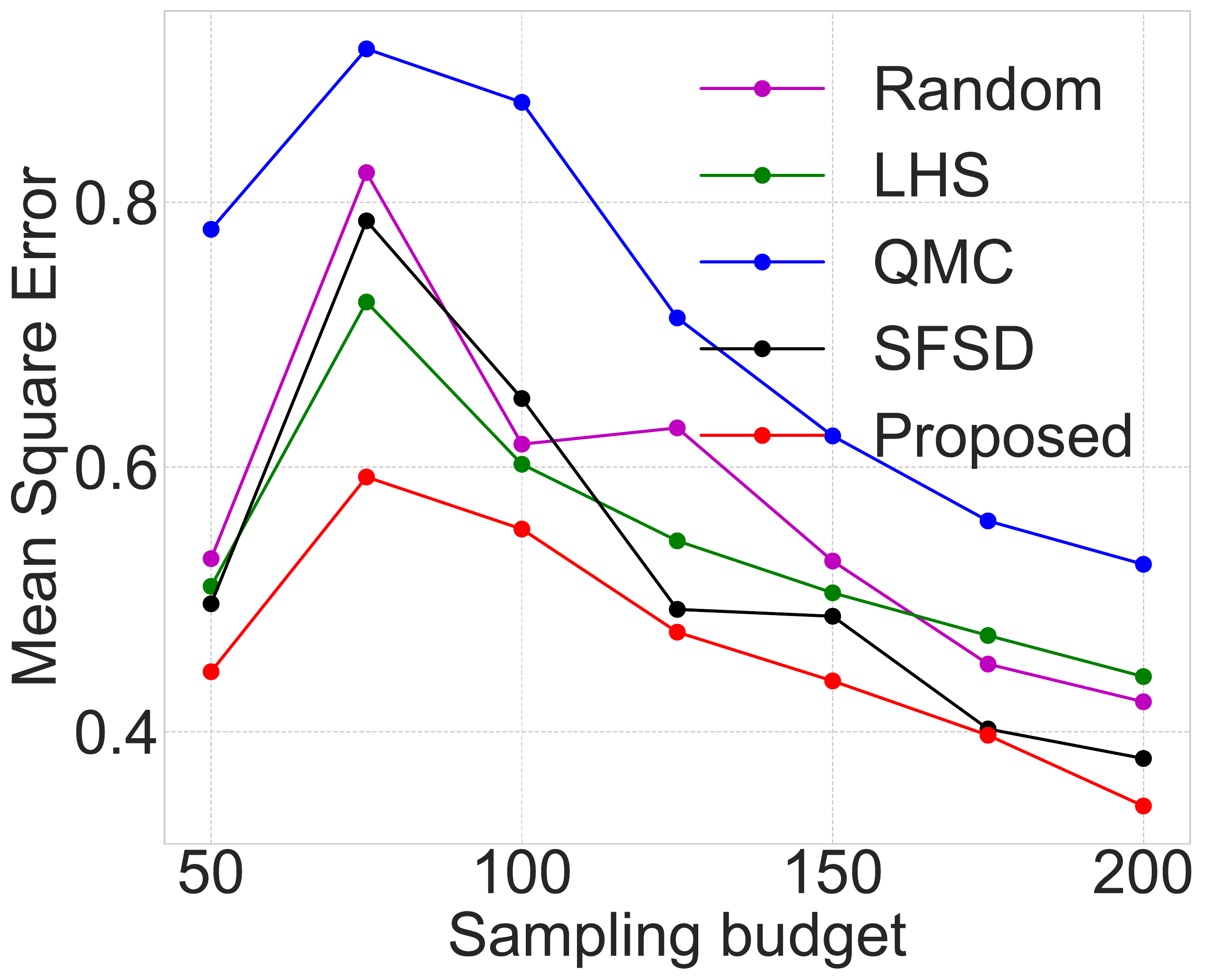} \label{fig:Bayes_ackley_5D} 
	}
	\caption{\textit{Sampling for Predictive Modeling}: Performance of sample designs in recovering regression functions using : (a)-(f) blind exploration, (g)-(l) sequential sampling.}
	
	\label{fig:DS_BO_results}
	\vspace{-0.1in}
\end{figure*}

\begin{table*}[t]
	\caption{\textit{Hyper-parameter search for a feature extractor-classifier pipeline}: Average \textit{f1}-score obtained over $10$ realizations of exploratory sample designs for the $20$-Newsgroup dataset. This pipeline used \textit{tf-idf} features and a \textit{SGD} classifier.}
	\centering
	\renewcommand*{\arraystretch}{1.3}
	\resizebox{0.8\textwidth}{!}{
		\begin{tabular}{ccccccc}
			\hline          
			\rowcolor{gray!50}\multicolumn{7}{l}{\bf Blind Exploration} \\
			\hline
			
			\cellcolor{gray!15}{\textbf{d}} & \cellcolor{gray!15}\textbf{N}   & \cellcolor{gray!15}\textbf{QMC}    & \cellcolor{gray!15}\textbf{LHS}             & \cellcolor{gray!15}\textbf{Random} & \cellcolor{gray!15}\textbf{SFSD} & \cellcolor{gray!15}\textbf{Proposed}        \\
			
			5 & 50  & {84.0995 $\pm$ 0}& 83.838 $\pm$ 0.415 & 83.872 $\pm$ 0.257 & 84.328 $\pm$ 0.166 & \textbf{84.441 $\pm$ 0.064}  \\
			5 & 75  & {84.1758 $\pm$ 0}& 84.179 $\pm$ 0.237 & 84.196 $\pm$ 0.172 & \textbf{84.431 $\pm$ 0.138}   & 84.414 $\pm$ 0.100 \\
			5 & 100 & {84.1543 $\pm$ 0}& 84.227 $\pm$ 0.189 & 84.161 $\pm$ 0.205 & 84.331 $\pm$ 0.127   & \textbf{84.463 $\pm$ 0.075}  \\
			5 & 125 & {84.3501 $\pm$ 0}& 84.292 $\pm$ 0.131 & 84.286 $\pm$ 0.158 & 84.458 $\pm$ 0.073   &  \textbf{84.482 $\pm$ 0.089}  \\
			5 & 150 & {84.3294 $\pm$ 0}& 84.314 $\pm$ 0.141 & 84.336 $\pm$ 0.185 & 84.453 $\pm$ 0.058   & \textbf{84.527 $\pm$ 0.028}  \\
			\hline
	\end{tabular}}
	\label{table:newsgroup}
\end{table*}

\begin{table*}[th]
	\caption{\textit{Hyper-parameter search to build deep networks for MNIST digit classification}: Best test accuracy obtained through the inclusion of hyper-parameter optimization using different sample designs. Note that, we consider both blind exploration and sequential sampling settings, and the results reported are averages over $10$ independent realizations of the sample design.}
	\centering
	\renewcommand*{\arraystretch}{1.3}
	\resizebox{0.8\textwidth}{!}{
		\begin{tabular}{ccccccc}
			\hline
			\rowcolor{gray!50}\multicolumn{7}{l}{\bf Blind Exploration} \\
			\hline
			
			\cellcolor{gray!15}{\textbf{d}} & \cellcolor{gray!15}\textbf{N}   & \cellcolor{gray!15}\textbf{QMC}    & \cellcolor{gray!15}\textbf{LHS}             & \cellcolor{gray!15}\textbf{Random} & \cellcolor{gray!15}\textbf{SFSD} & \cellcolor{gray!15}\textbf{Proposed}        \\
			
			3 & 50  & 98.57 $\pm$ 0  & {91.198 $\pm$ 3.492} & 98.554 $\pm$ 0.229 & 98.691 $\pm$ 0.316 & \textbf{98.79 $\pm$ 0.198}  \\
			3 & 100 & 98.82 $\pm$ 0 & 98.794 $\pm$ 0.171 & 98.688 $\pm$ 0.431 & {98.873 $\pm$ 0.124} & \textbf{98.896 $\pm$ 0.098} \\
			3 & 200 & 98.92 $\pm$ 0 & 98.932 $\pm$ 0.033 & 98.921 $\pm$ 0.126 & \textbf{98.975 $\pm$ 0.125} & {98.969 $\pm$ 0.035}  \\
			4 & 50  & 98.66 $\pm$ 0 & 98.116 $\pm$ 0.690 & 97.818 $\pm$ 1.285 & 98.623 $\pm$ 0.325 & \textbf{98.806 $\pm$ 0.138} \\
			4 & 100 & 98.61 $\pm$ 0 & 98.70 $\pm$ 0.215 & 98.654 $\pm$ 0.216 & 98.748 $\pm$ 0.202 & \textbf{98.976 $\pm$ 0.157} \\
			4 & 200 & 98.199 $\pm$ 0 & 98.832 $\pm$ 0.117 & 98.902 $\pm$ 0.134 & {98.921 $\pm$ 0.075} & \textbf{98.932 $\pm$ 0.061} \\
			5 & 50  & 98.188 $\pm$ 0 & 97.996 $\pm$ 0.737 & {91.06 $\pm$ 3.485}  & 98.622 $\pm$ 0.238 & \textbf{98.832 $\pm$ 0.134} \\
			5 & 100 & 98.77 $\pm$ 0 & 98.802 $\pm$ 0.163 & 98.642 $\pm$ 0.233 & \textbf{98.846 $\pm$ 0.188} & {98.834 $\pm$ 0.148} \\
			5 & 200 & 98.73 $\pm$ 0 & \textbf{98.992 $\pm$ 0.102} & 98.862 $\pm$ 0.131 & 98.944 $\pm$ 0.118 & 98.967 $\pm$ 0.068\\
			\hline        
			
			\rowcolor{gray!50}\multicolumn{7}{l}{\bf Sequential Sampling} \\
			\hline
			\cellcolor{gray!15}{\textbf{d}} & \cellcolor{gray!15}\textbf{N}   & \cellcolor{gray!15}\textbf{QMC}    & \cellcolor{gray!15}\textbf{LHS}             & \cellcolor{gray!15}\textbf{Random} & \cellcolor{gray!15}\textbf{SFSD} & \cellcolor{gray!15}\textbf{Proposed}        \\
			
			3   & 100 & 97.354 $\pm$ 0   & 97.466 $\pm$ 0.081 & 97.49 $\pm$ 0.445  & 97.367 $\pm$ 0.371 & \textbf{97.626 $\pm$ 0.128 }  \\
			4   & 100 & 97.581 $\pm$ 0  & 96.952 $\pm$ 0.562  & 97.492 $\pm$ 0.135 & \textbf{97.628 $\pm$ 0.198 } & {97.597}$\pm$ 0.196  \\
			5   & 100 & 94.222 $\pm$ 0  & 97.296 $\pm$ 0.434  & 96.171 $\pm$ 0.950 & 97.487 $\pm$ 0.3134  & \textbf{97.662 $\pm$ 0.110 } \\
			
			\hline
			
	\end{tabular}}
	\label{table:mnist}
\end{table*}

In this section, we study the qualitative performance of the proposed coverage-based design in predictive modeling, where one needs to recover unknown regression functions using a given set of sample observations. The goal of this study is to understand the impact of the improved coverage properties in the proposed design and the effectiveness of our sample synthesis algorithm. We consider both blind exploration, where the model is constructed only using one-shot exploratory samples, and sequential sampling, where the exploration samples are used to initialize a Bayesian optimization (Bayes-Opt) pipeline. Bayes-Opt~\cite{snoek2012bayesianopt} is a widely adopted sequential design framework typically employed for global optimization of complex functions. These methods begin by constructing a surrogate for the unknown function based on an initial sample, and then sequentially allocate the remaining design budget to quantify uncertainties of the surrogate, and utilize an acquisition function (e.g. expected improvement) to choose the next sample. We present comparisons to popular sampling methods, namely uniform random, Latin Hypercube sampling (LHS), Sobol (QMC) sequences, and SFSD, which is a state-of-the-art coverage-based design technique \cite{kailkhura2018spectral}. We show that the proposed approach produces superior recovery performance, thus establishing coverage-based designs as an effective solution for exploratory sampling.

\noindent \textbf{Setup}: We use the following benchmark functions from the global optimization literature: Alpine $N.1$ and Ackley in dimensions $3$, $4$ and $5$, respectively. In order to evaluate the generalization of fitted functions, we generate $10^4$ test samples using a regular grid in the sampling region, and use the mean squared error (MSE) with respect to the true function as the evaluation metric. For all experiments, we used random forest regressors with $100$ trees, and the results reported were obtained by averaging over $20$ independent realizations of sample designs.

\noindent \textbf{Blind exploration}: Figure \ref{fig:DS_BO_results} (a)-(f) compare the performance of our approach to the baseline methods in the fully exploratory case. It can be seen that the proposed design consistently outperforms popularly adopted sampling methods across varying $N$ ($50$ to $200$). Another striking observation is that there is significant variability in performance of the widely-adopted QMC sequences across dimensions, and as $d$ increases it can sometimes perform even worse than uniform random samples. Furthermore, the poor performance of models learned using LHS and uniform random sampling for $d>3$ can be directly attributed to their poor space-filling properties. Although SFSD and the proposed approach belong to the family of coverage-based designs, due to the improved coverage characteristics, our method consistently outperforms SFSD in all cases. 

\noindent \textbf{Sequential sampling}: In this experiment, we study the impact of the choice for initial design on a Bayes-Opt pipeline. We consider an initial sampling budget of $N=50$, and then sequentially sample $150$ more samples to evaluate the behavior of Bayes-Opt on the same set of functions used in the previous case. Similar to the blind exploration case, we observe in Figure \ref{fig:DS_BO_results} (g)-(l) that the proposed design performs significantly better compared to other state-of-the-practice choices. Although QMC sequences perform reasonably better than uniform random at $d=3$, their performance degrades as $d$ grows.


\section{Application: Hyper-Parameter Search}
\begin{figure*}[h] 
	\centering
	\subfigure[$d=3, N=50$]{%
		\includegraphics[width=.3\textwidth]{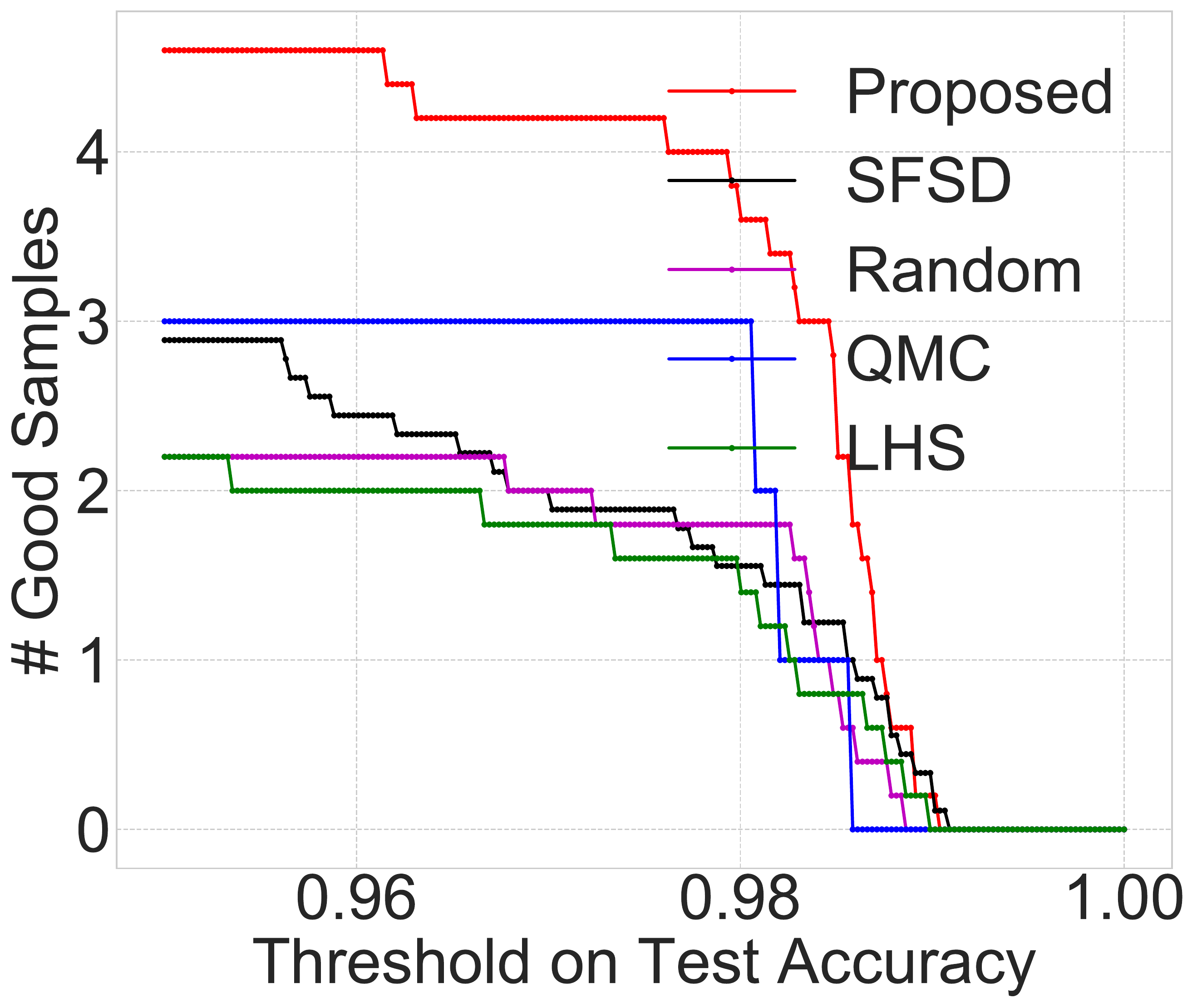} \label{fig:blindhopt_50_3D}
	}
	\subfigure[$d=3, N=100$]{%
		\includegraphics[width=.3\textwidth]{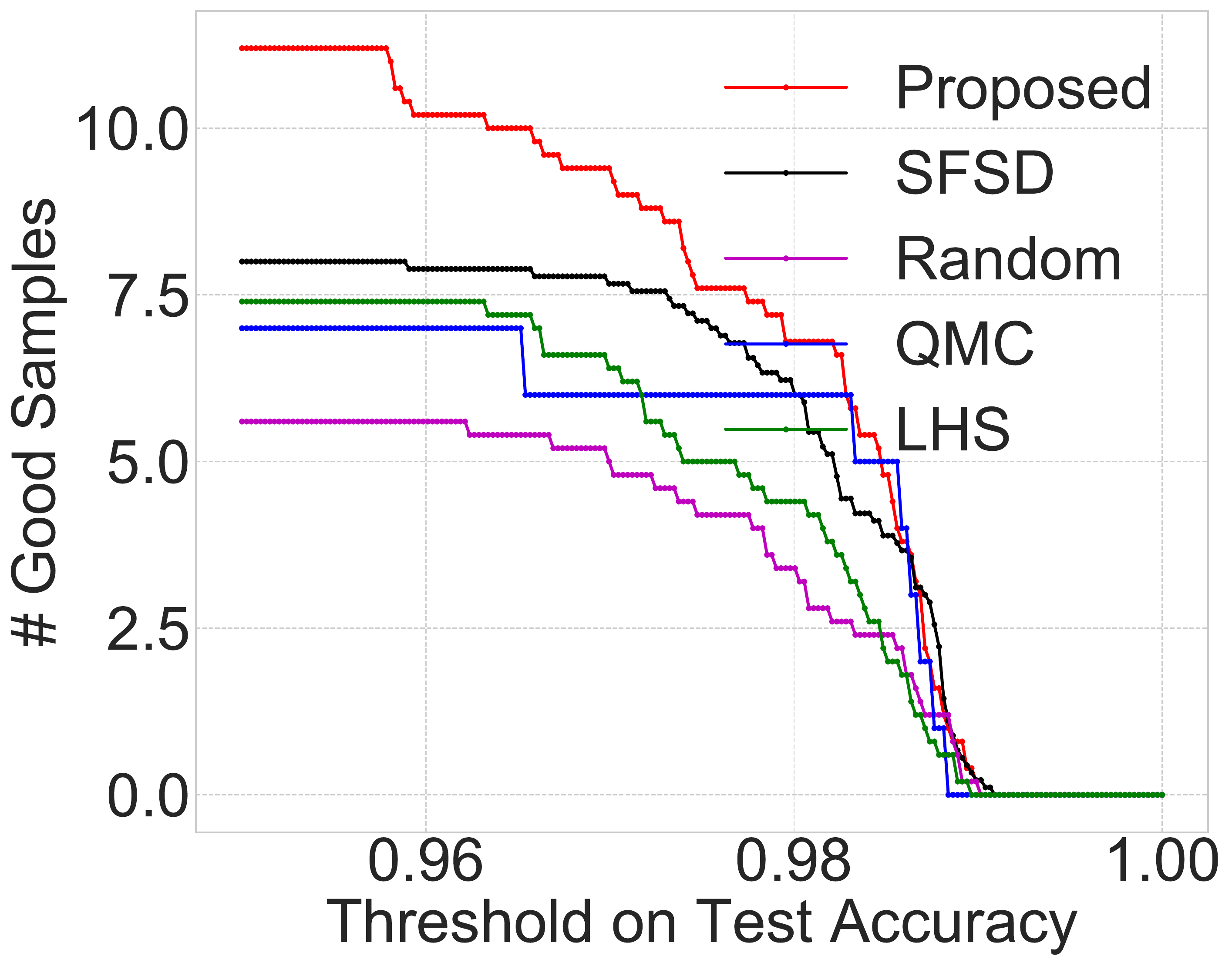} \label{fig:blindhopt_100_3D} 
	} 
	\subfigure[$d=3, N=200$]{%
		\includegraphics[width=.3\textwidth]{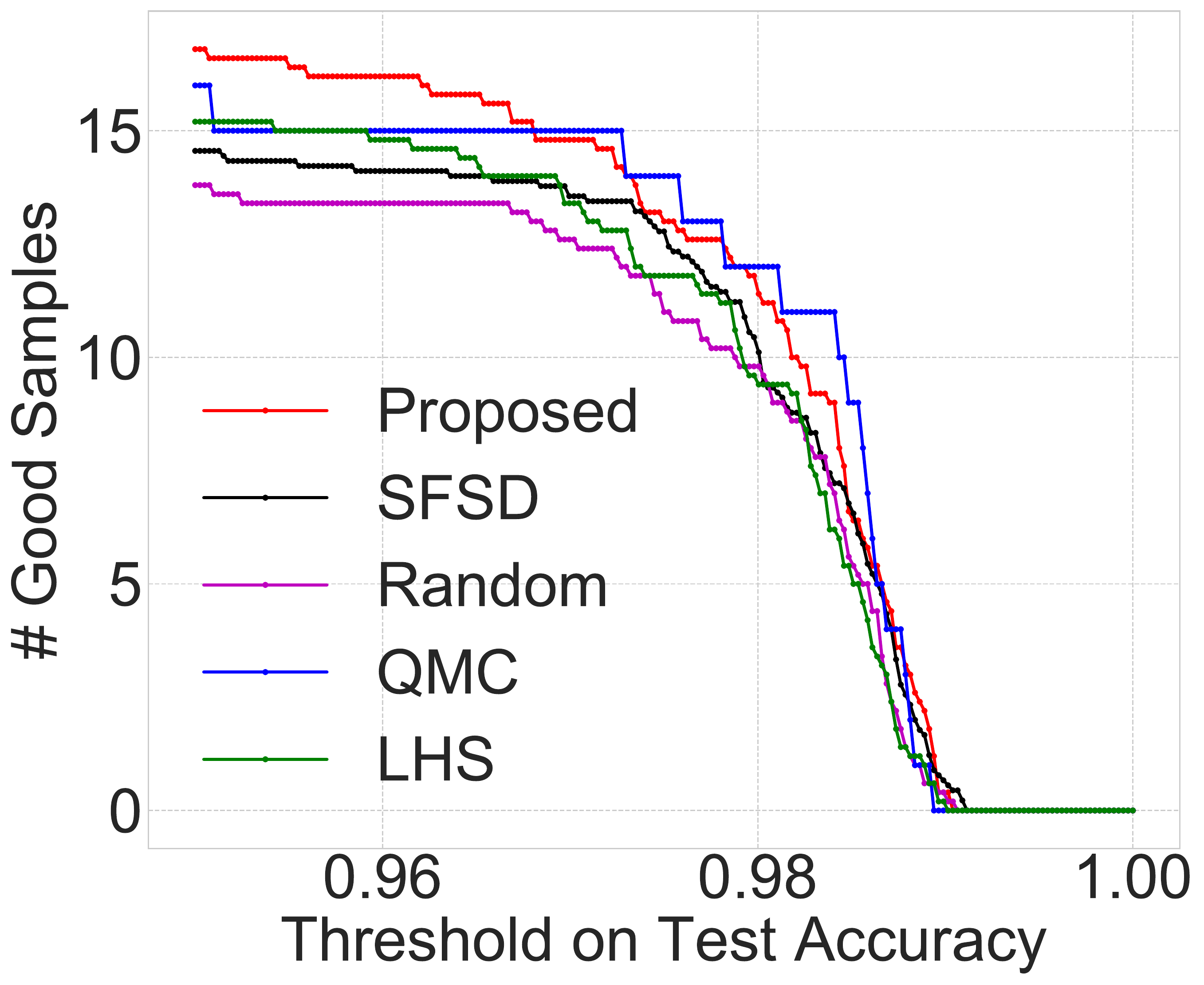} \label{fig:blindhopt_200_3D}
	}
	\subfigure[$d=4, N=50$]{%
		\includegraphics[width=.3\textwidth]{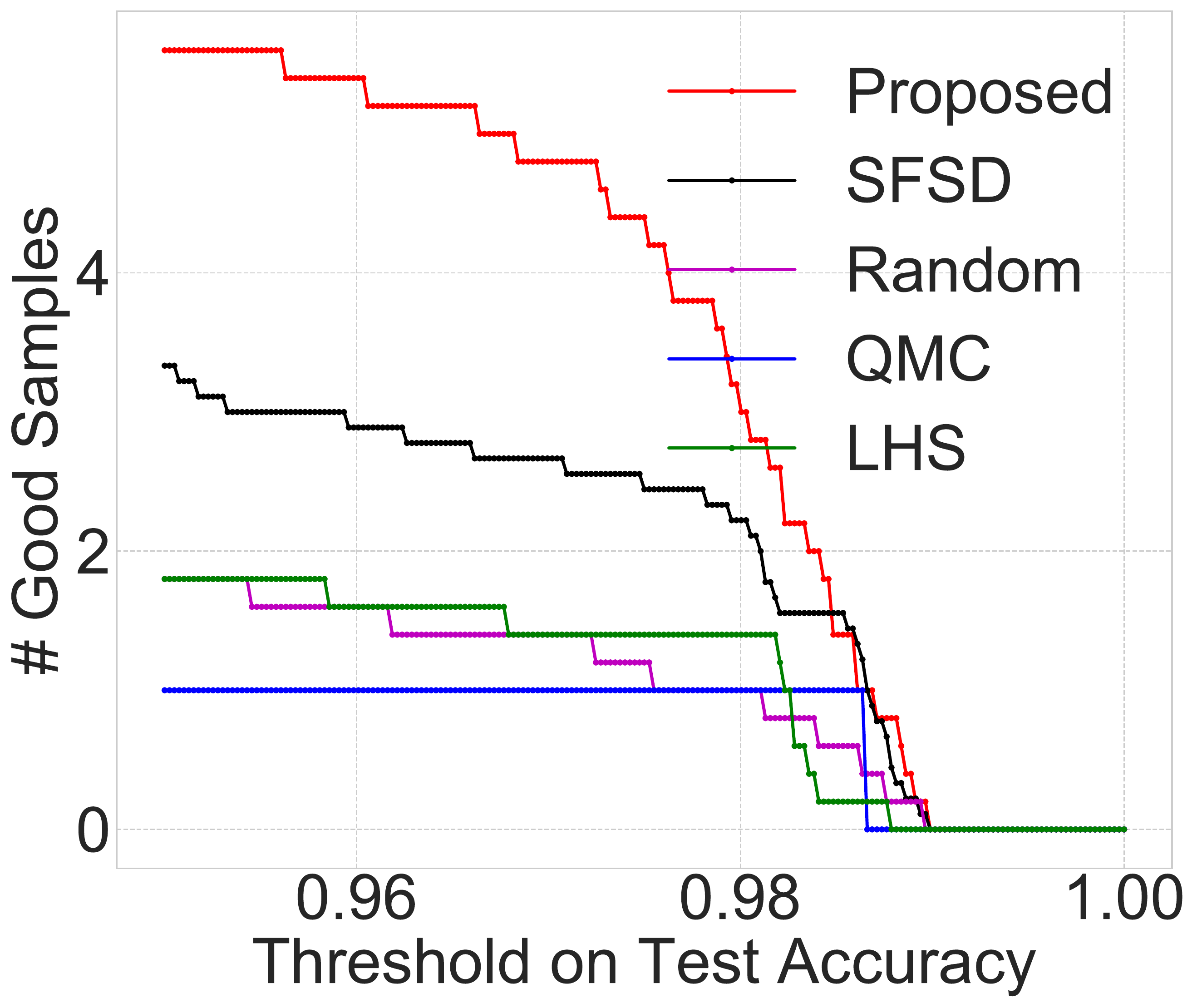} \label{fig:blindhopt_50_4D} 
	} 
	\qquad
	\subfigure[$d=4, N=100$]{%
		\includegraphics[width=.3\textwidth]{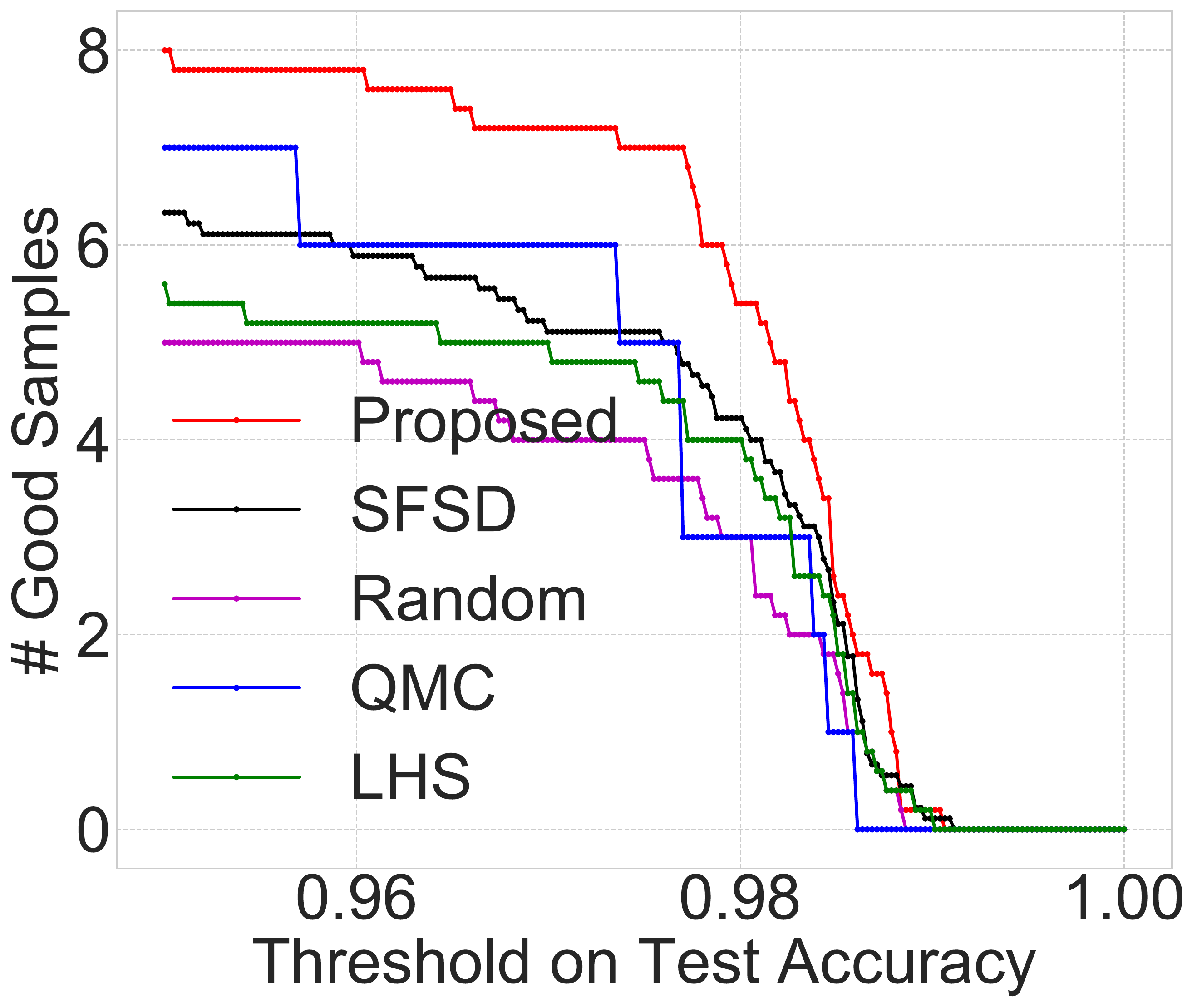} \label{fig:blindhopt_100_4D}
	}
	\subfigure[$d=4, N=200$]{%
		\includegraphics[width=.3\textwidth]{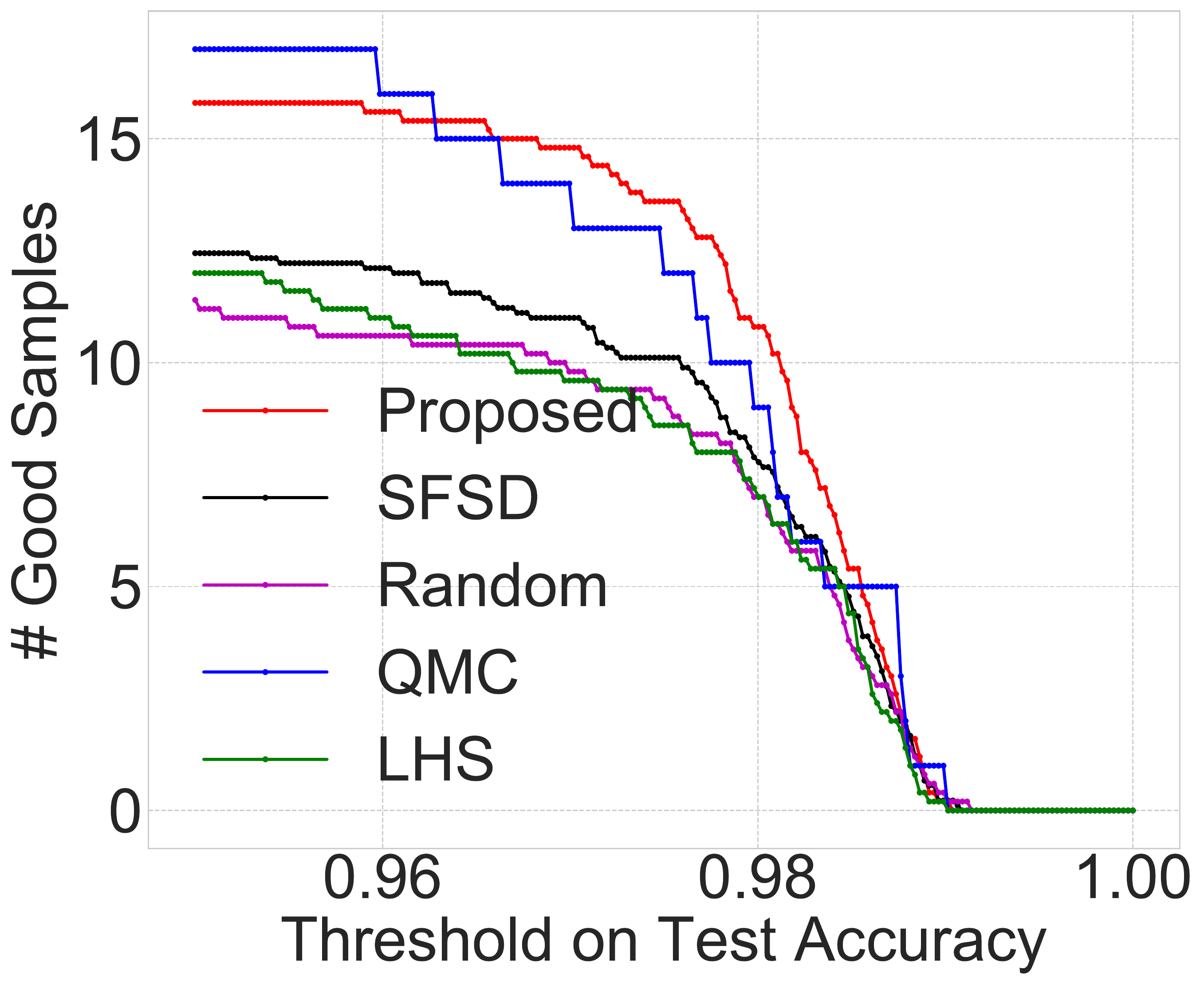} \label{fig:blindhopt_200_4D} 
	} 
	\subfigure[$d=5, N=50$]{%
		\includegraphics[width=.3\textwidth]{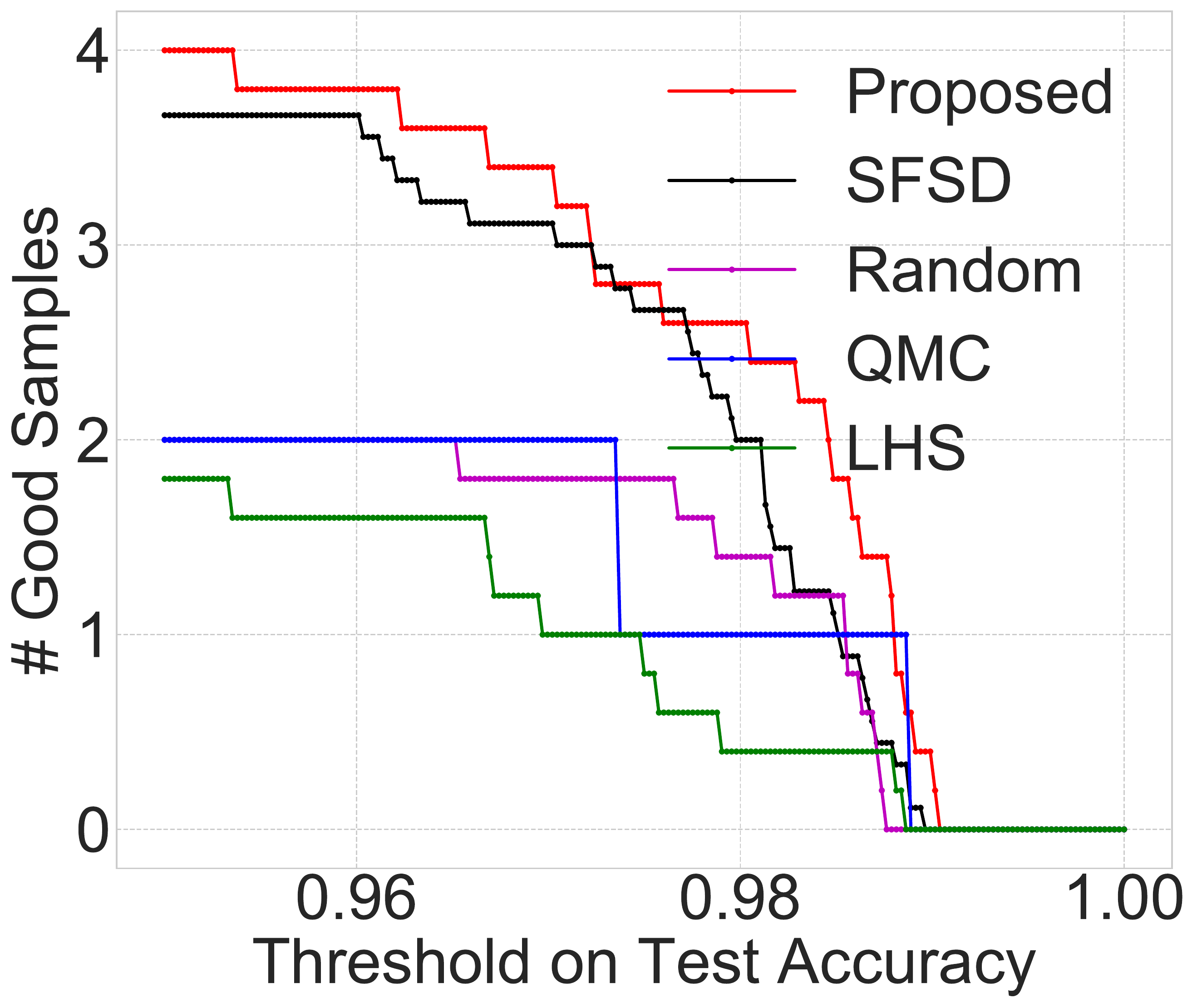} \label{fig:blindhopt_50_5D} 
	} 
	\subfigure[$d=5, N=100$]{%
		\includegraphics[width=.3\textwidth]{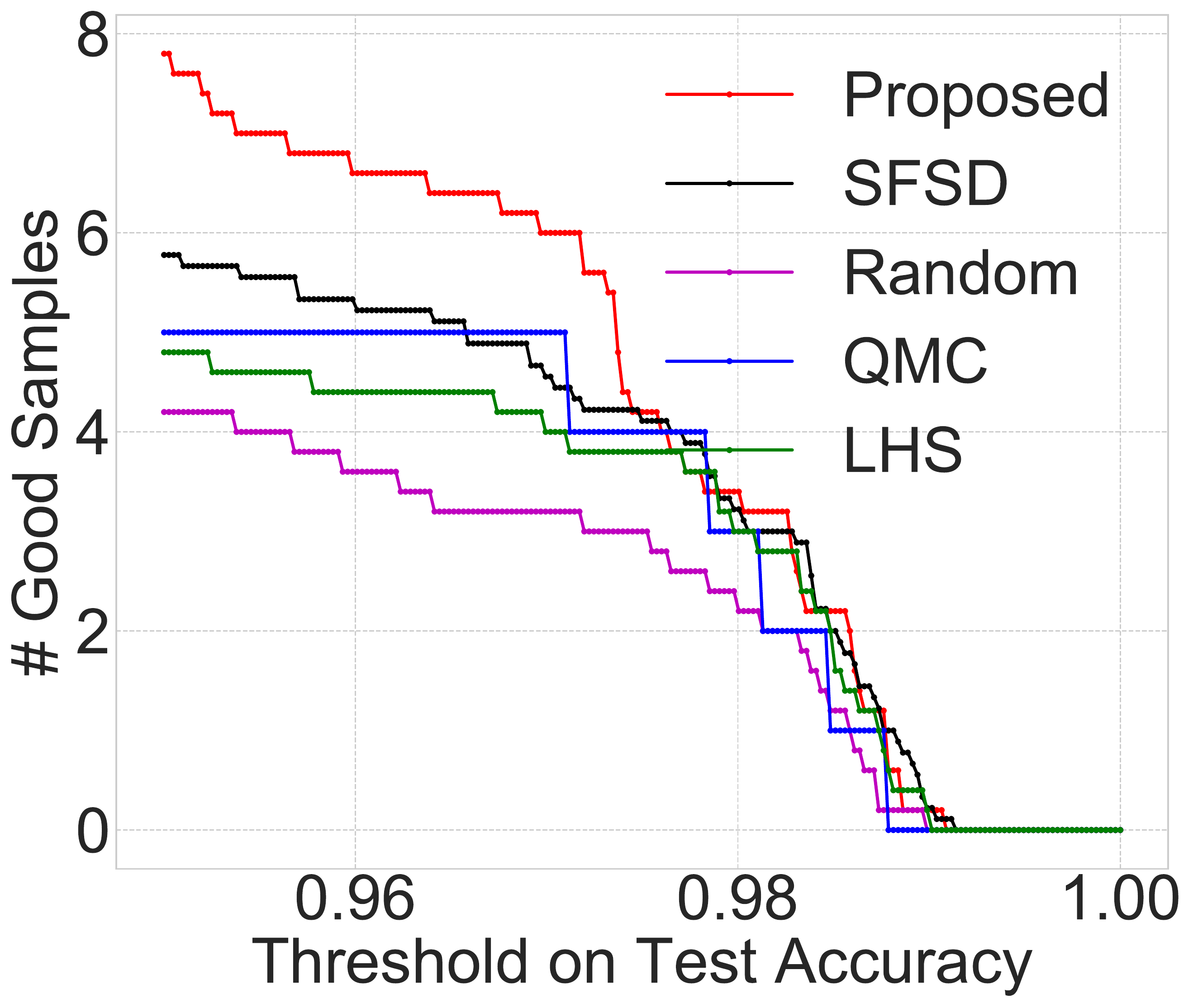} \label{fig:blindhopt_100_5D}
	}
	\subfigure[$d=5, N=200$]{%
		\includegraphics[width=.3\textwidth]{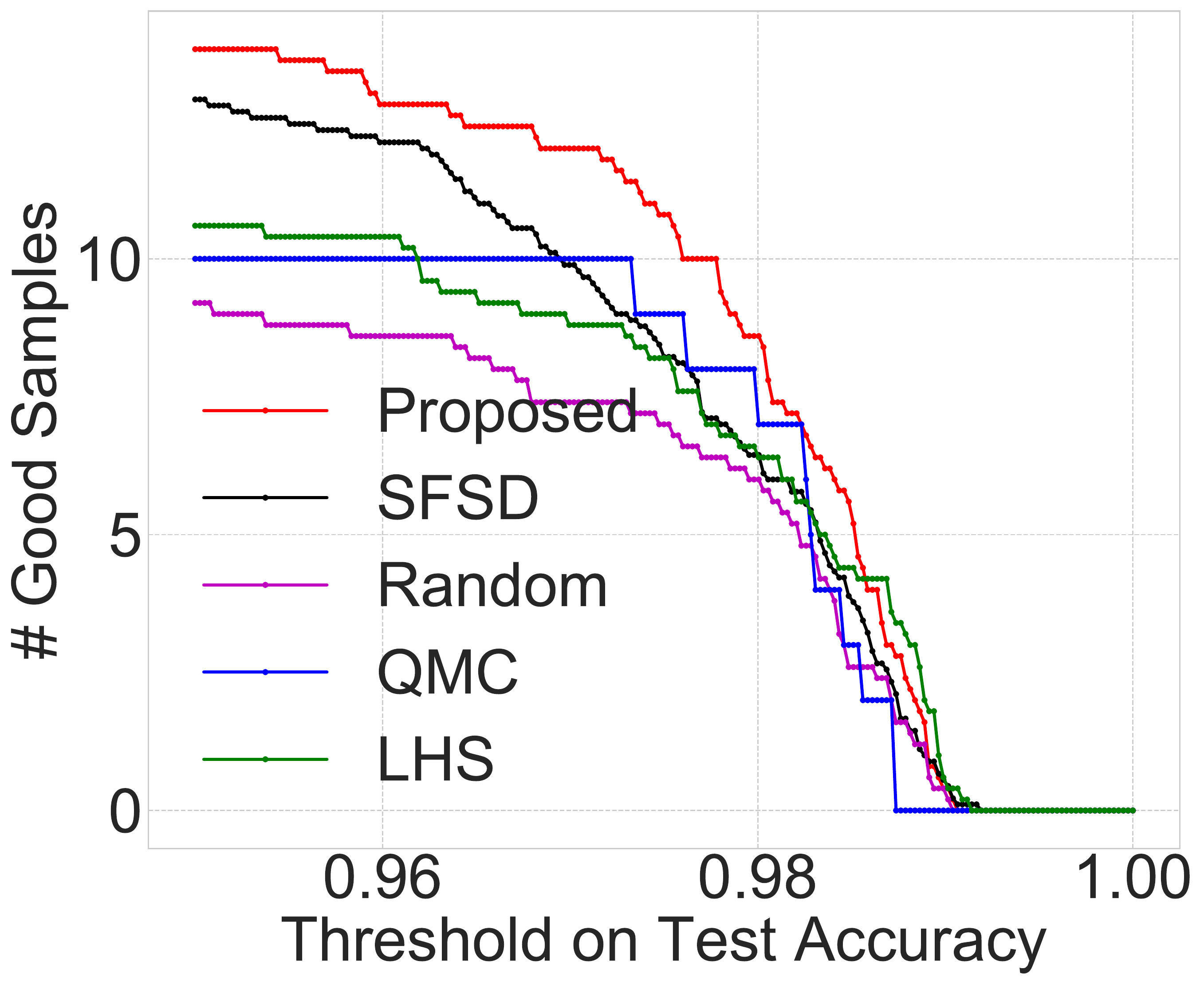} \label{fig:blindhopt_200_5D}
	}
	\caption{\textit{Hyper-parameter search to build deep networks MNIST digit recognition}: Precision metric obtained through blind exploration with different sample designs.} 
	\label{fig:MS_results}
\end{figure*}

\begin{figure*}[th] 
	\centering
	\subfigure[$d=3$]{%
		\includegraphics[width=.3\textwidth]{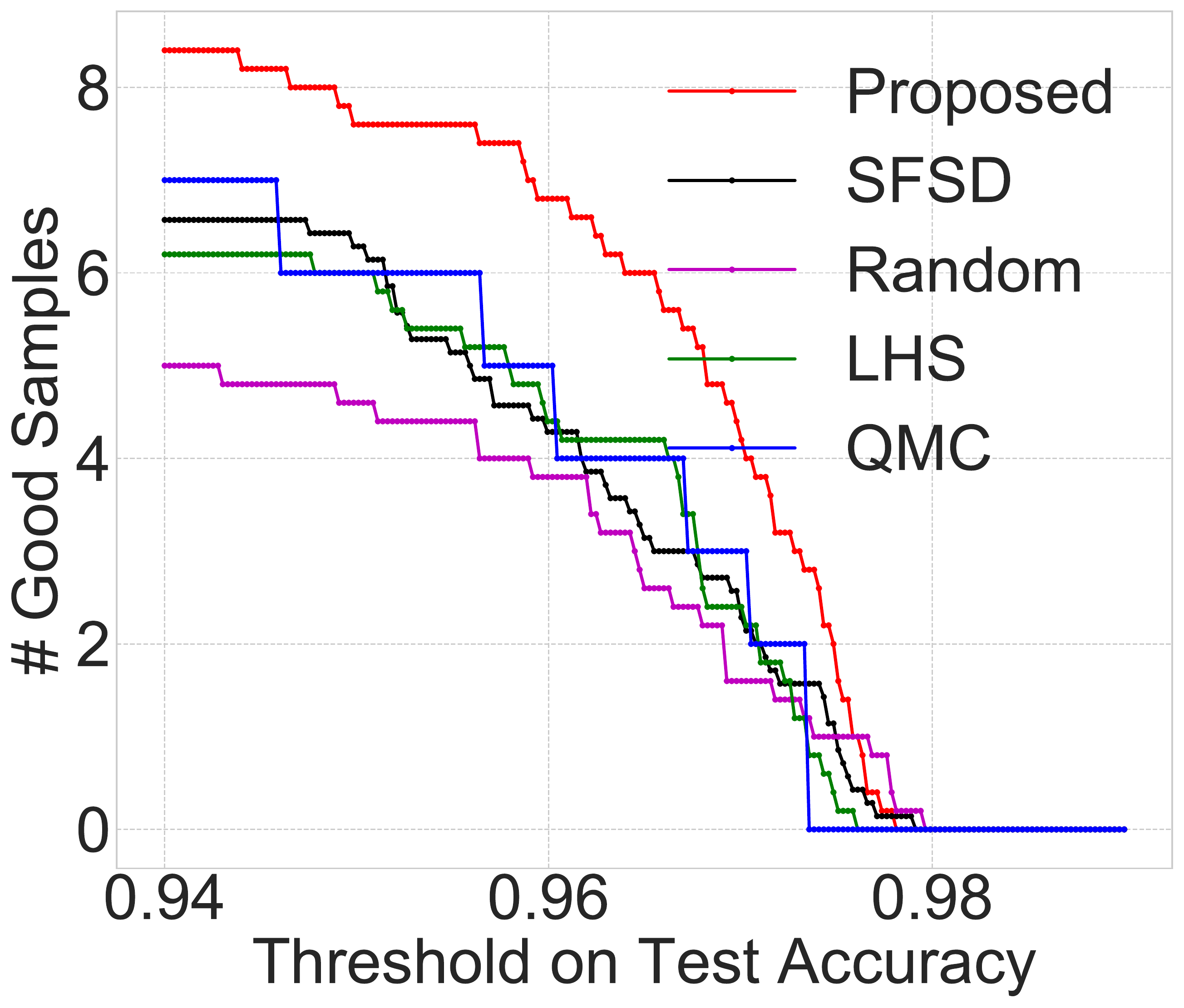} \label{fig:bayesopt_50_3D}
	}
	\subfigure[$d=4$]{%
		\includegraphics[width=.3\textwidth]{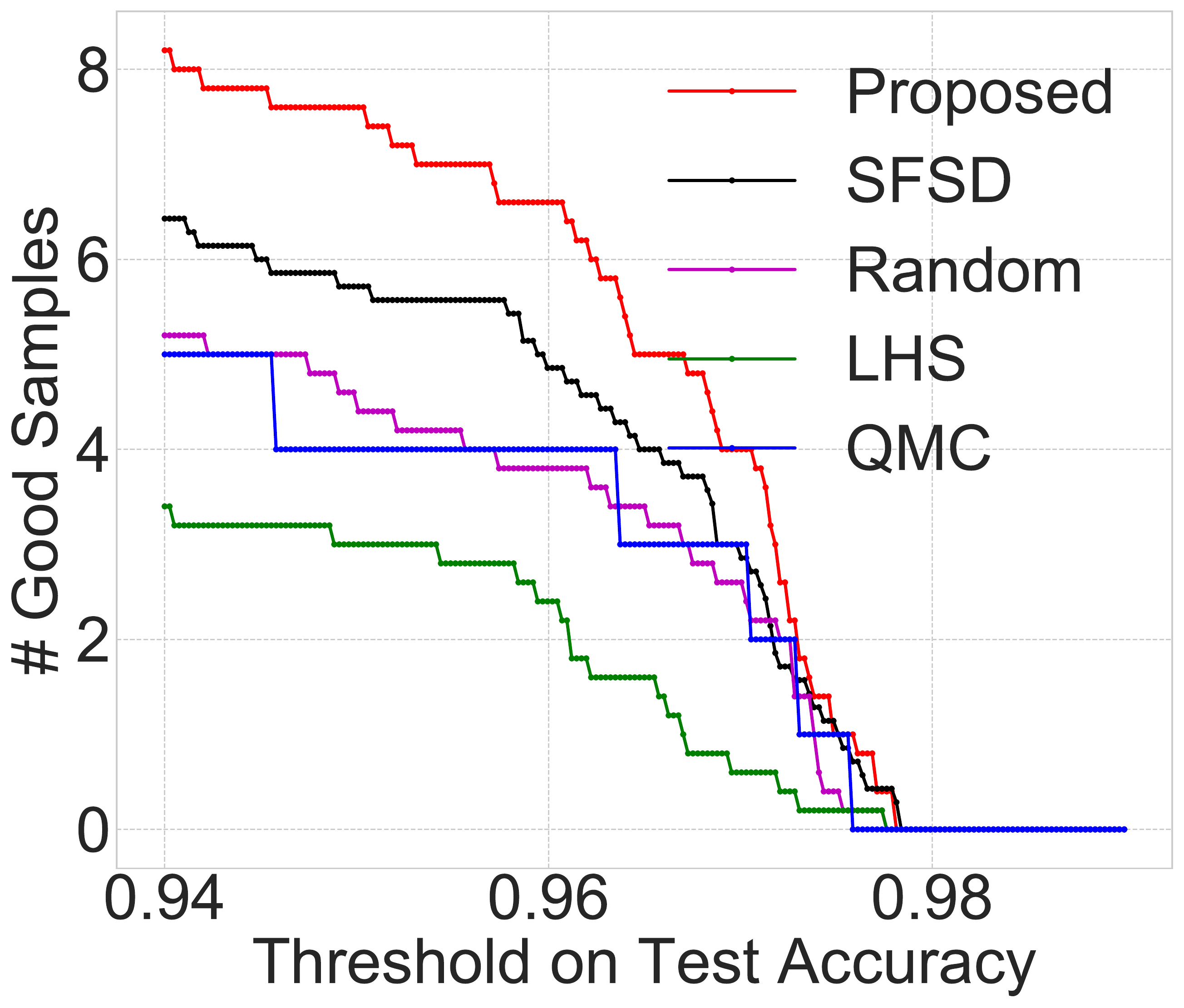} \label{fig:bayesopt_100_3D} 
	} 
	\subfigure[$d=5$]{%
		\includegraphics[width=.3\textwidth]{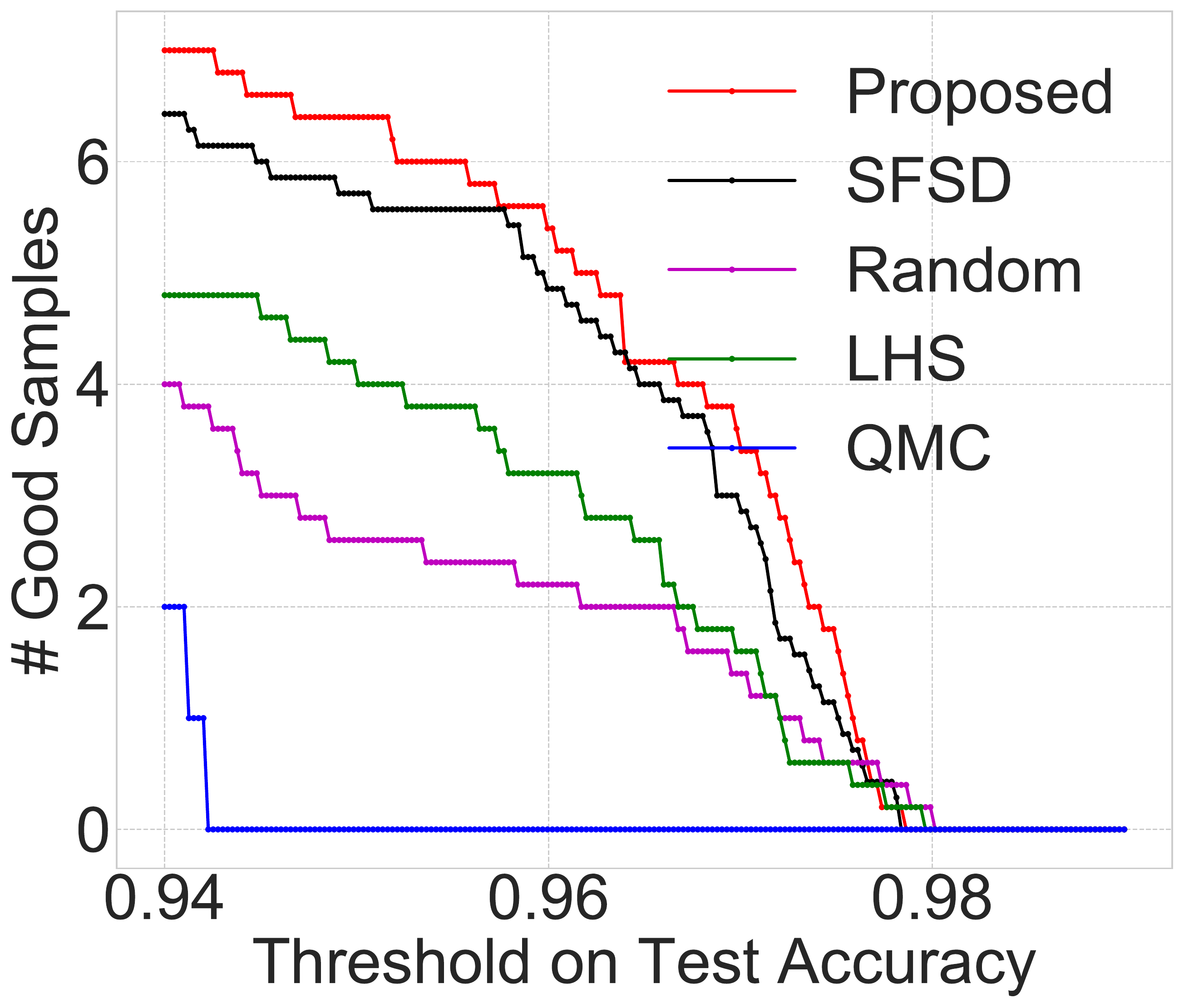} \label{fig:bayesopt_200_3D} 
	}    
	\caption{\textit{Hyper-parameter search to build deep networks MNIST digit recognition}: Precision metric obtained through Bayes-Opt with different initial exploratory samples.} 
	\label{fig:MS_BO_results}
\end{figure*}

Hyper-parameter search is critical to modern machine learning algorithms and resource-efficient optimization is directly linked to the scalability of the solutions. For this experiment, we consider both a conventional feature extractor-classifier pipeline and end-to-end deep learning systems, where the goal is to minimize the validation error~\cite{bergstra2011algorithms}. The search space is characterized by a sparse set of locally optimal solutions, and requires effective sampling to rapidly choose a well-performing configuration. The evaluation metric that we use is the \textit{precision}, i.e., the number of selected configurations that produces validation accuracies greater than a pre-defined threshold $\tau$. We use this proxy metric~\cite{bergstra2011algorithms,bousquet2017critical} since the global optimum is unknown, and more importantly identifying multiple locally optimal configurations in the search space reflects the ability of a sampling technique in characterizing the response surface. For completeness, we also include the widely-used \textit{best validation accuracy} achieved over multiple realizations of the considered sample designs. 


\subsection{Conventional Feature Extractor-Classifier Pipeline}
In this experiment, we consider the problem of choosing hyper-parameters for feature extraction and classification of text documents in the \textit{20-Newsgroup} dataset. This is a collection of approximately $20,000$ documents, partitioned evenly across $20$ different newsgroups. We use a feature extractor-classifier pipeline that consists of : (a) \textit{Count Vectorizer}: converts a collection of text documents into a matrix of token counts; (b) \textit{Tf-idf-Transformer}: transforms a count matrix into a normalized (term-frequency times inverse document-frequency) tf-idf representation; and (c) linear classifier with stochastic gradient descent (SGD) training. We considered $5$ hyper-parameters -- document frequency threshold and maximum number of features in the feature extraction step, and $3$ settings for classifier design : number of iterations, learning rate and regularization penalty.

We vary the sampling budget $N$ in the range $(50,150)$ and compute the $f1$-score (macro-averaged) from the best performing configuration in the exploratory sample. We report the mean and standard deviation obtained using $10$ independent realizations of the samples. As showed in Table~\ref{table:newsgroup}, in most of the cases, coverage-based designs outperform other random sampling baselines both in terms of expected performance and variance. More specifically, the proposed approach identifies the best configuration in every case. The superior performance of the proposed design over SFSD can be attributed to the improved coverage in the synthesized samples. On the other hand, conventional methods such as \textit{LHS} and \textit{Random} suffer from high variance across realizations.




\subsection{Building Deep Models for MNIST Digit Classification}
In this section, we consider the problem of building deep networks for classifying handwritten digits from MNIST, which contains $50,000$ train and $10,000$ test images. We evaluate the proposed sample design approach under blind exploration and sequential sampling settings.

\noindent \textbf{Blind exploration}: We use a simple CNN architecture: $
 conv[3 \times 3 \times 8] \to conv[3 \times 3 \times 16] \to FC[128] \to FC[64] \to FC[10].
$with \rm{ReLU} activation and dropouts in after every layer. The training was carried out using gradient descent with the momentum optimizer. The set of $5$ hyper-parameters included learning rate, momentum and dropouts at the $2^{nd}$ $conv$ layer, the $1^{st}$ $FC$ layer and the $2^{nd}$ $FC$ layer respectively. We also considered $4-d$ and $3-d$ subsets, where some of the dropouts fixed at $0.5$. For this experiment, we used the sampling budgets $N = \{50, 100 ,200\}$. In each case, we estimated the \textit{precision} metric by varying the threshold $\tau$, between $0.95$ and $1$. All results reported were averaged over $10$ independent realizations.

Figure \ref{fig:MS_results} and Table~\ref{table:mnist} illustrate the validation performance of different sample designs for this problem. We observe that the proposed design consistently achieves superior precision over existing experimental designs, thus ensuring a high probability of obtaining a generalizable model, particularly at lower sampling budgets. Although LHS and QMC samples perform reasonably well in some cases, their performance degrades as $d$ grows. Through the improved coverage characteristics, our approach sometimes identifies even twice as many local optima (based on the precision metric), thus motivating its use as an initializer for subsequent exploitation using Bayesian optimization. The results in Table~\ref{table:mnist} show that our method is able to sample the region of maximum interest consistently with low variance. Note that, the state-of-the-art SFSD design also performs consistently better than \textit{Random} and discrepancy based designs in terms of test accuracy, but often demonstrates a larger variance.

\begin{table*}[th]
	\caption{\textit{Hyper-parameter search to build CNNs for Cifar-10 image classification}: Best test accuracy obtained through the inclusion of hyper-parameter optimization using different sample designs. Note that, we consider both blind exploration and sequential sampling settings, and the results reported are averages over $10$ independent realizations of the sample design.}
	\centering
	\renewcommand*{\arraystretch}{1.3}
	\resizebox{0.8\textwidth}{!}{
		\begin{tabular}{ccccccc}
			\hline
			\rowcolor{gray!50}\multicolumn{7}{l}{\bf Blind Exploration} \\
			\hline
			\cellcolor{gray!15}{\textbf{d}} & \cellcolor{gray!15}\textbf{N}   & \cellcolor{gray!15}\textbf{QMC}    & \cellcolor{gray!15}\textbf{LHS}             & \cellcolor{gray!15}\textbf{Random} & \cellcolor{gray!15}\textbf{SFSD} & \cellcolor{gray!15}\textbf{Proposed} \\
			
			5 & 50 &  {80.36 $\pm$ 0}    & 80.023 $\pm$ 0.393   & 80.217 $\pm$ 0.217 & 80.145 $\pm$ 0.409 & \textbf{80.522 $\pm$ 0.334} \\
			5 & 100 & {80.70 $\pm$ 0} & 80.338 $\pm$ 0.245 & 80.448 $\pm$ 0.236 & 80.488 $\pm$ 0.136 & \textbf{80.959 $\pm$ 0.374} \\
			
			\hline        
			\rowcolor{gray!50}\multicolumn{7}{l}{\bf Sequential Sampling} \\
			\hline
			\cellcolor{gray!15}{\textbf{d}} & \cellcolor{gray!15}\textbf{N}   & \cellcolor{gray!15}\textbf{QMC}    & \cellcolor{gray!15}\textbf{LHS}             & \cellcolor{gray!15}\textbf{Random} & \cellcolor{gray!15}\textbf{SFSD} & \cellcolor{gray!15}\textbf{Proposed}        \\
			
			5& 100 & {80.842 $\pm$ 0.0} & 80.436 $\pm$ 0.206 & 80.623 $\pm$ 0.561 & 80.866 $\pm$ 0.341 & \textbf{81.033 $\pm$ 0.294} \\
			
			\hline                                      
	\end{tabular}}
	\label{table:cifar}
\end{table*}
\begin{figure*}[h] 
	\centering
	\subfigure[Blind exploration ($d=5,N=50$)]{%
		\includegraphics[width=.3\textwidth]{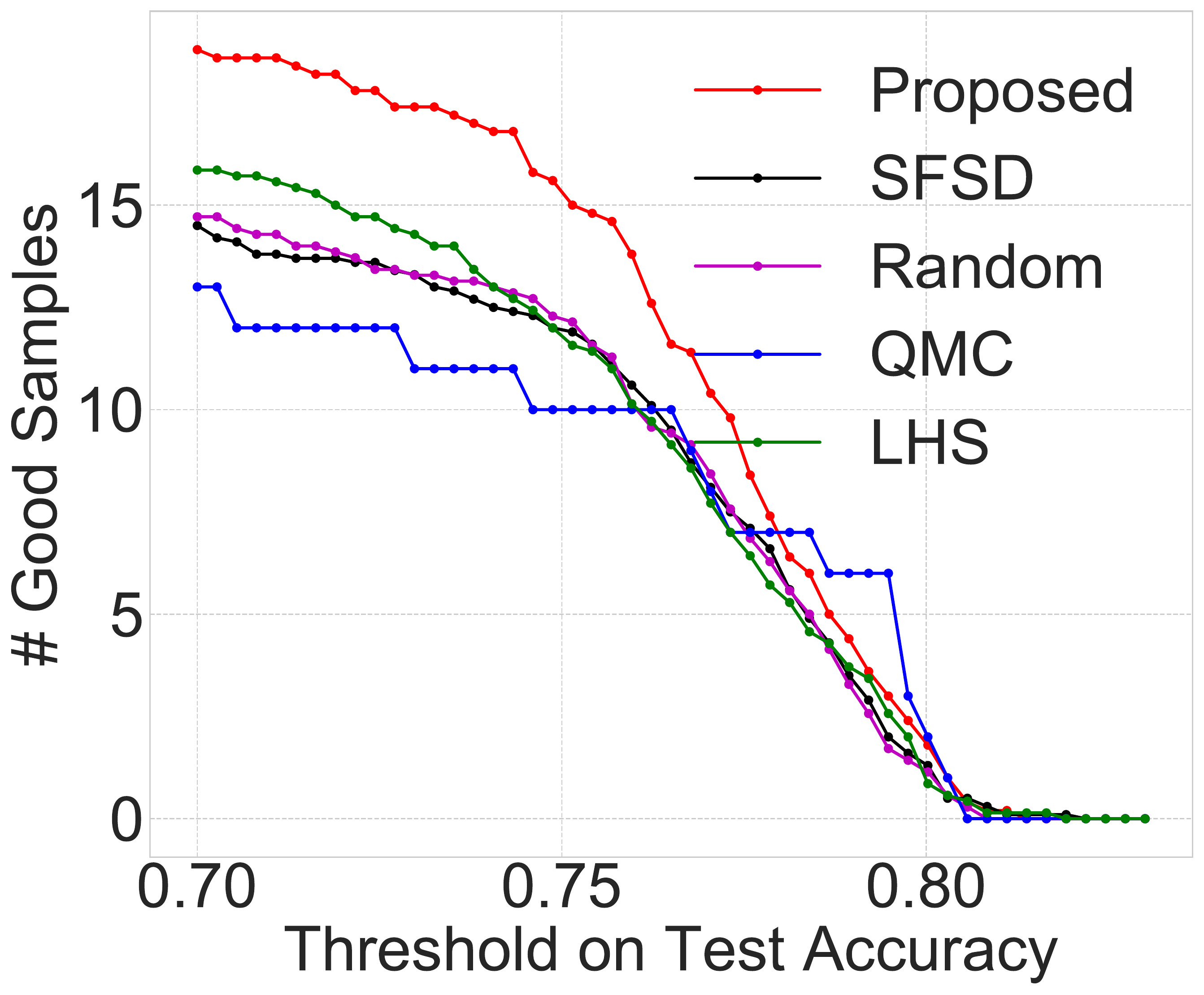} \label{fig:cif_blind_hopt_50_5D}
	}
	\subfigure[Blind exploration ($d=5,N=100$)]{%
		\includegraphics[width=.3\textwidth]{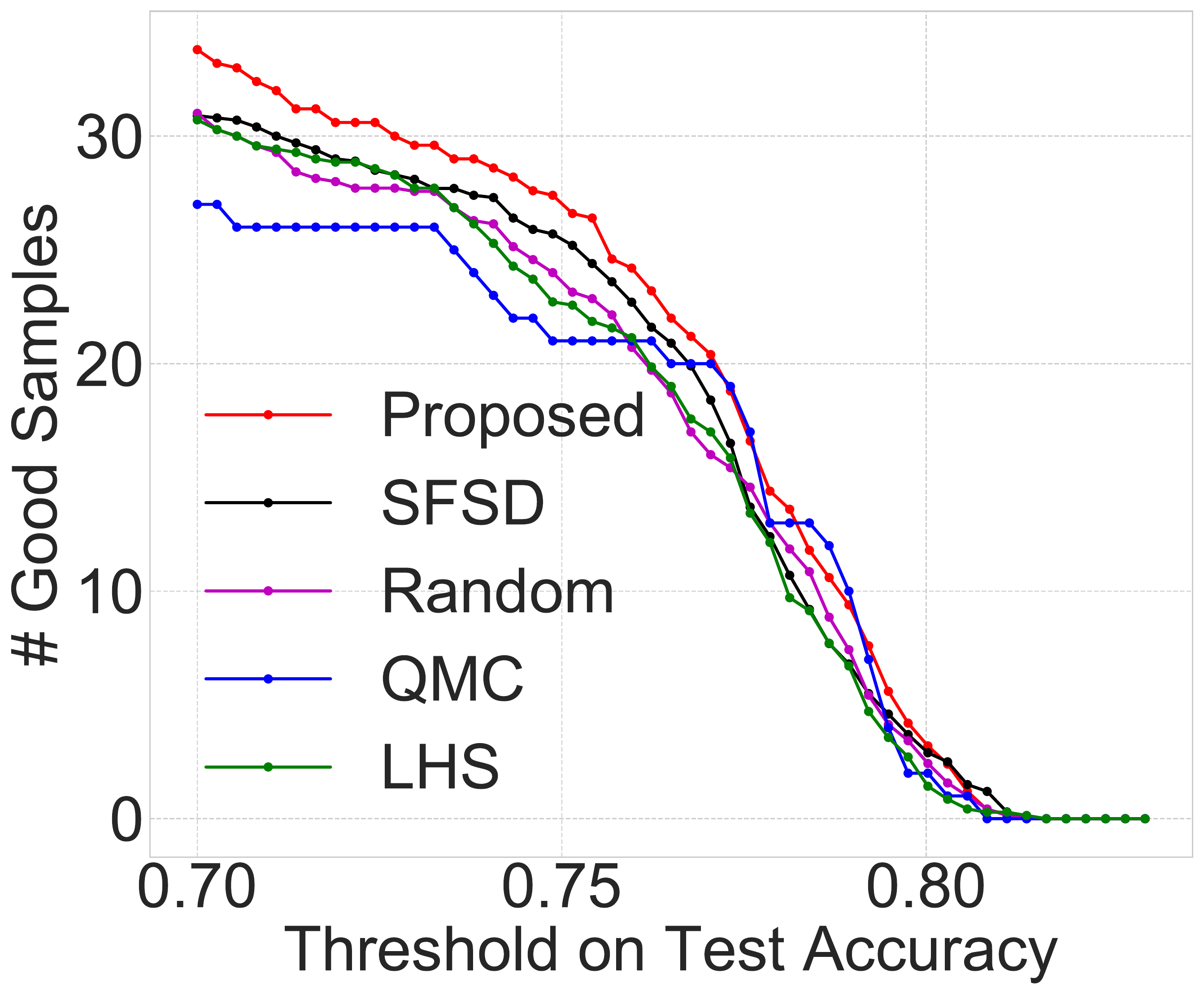} \label{fig:cif_blind_hopt_100_5D} 
	} 
	\subfigure[Sequential Sampling ($d=5, N = 100$)]{%
		\includegraphics[width=.3\textwidth]{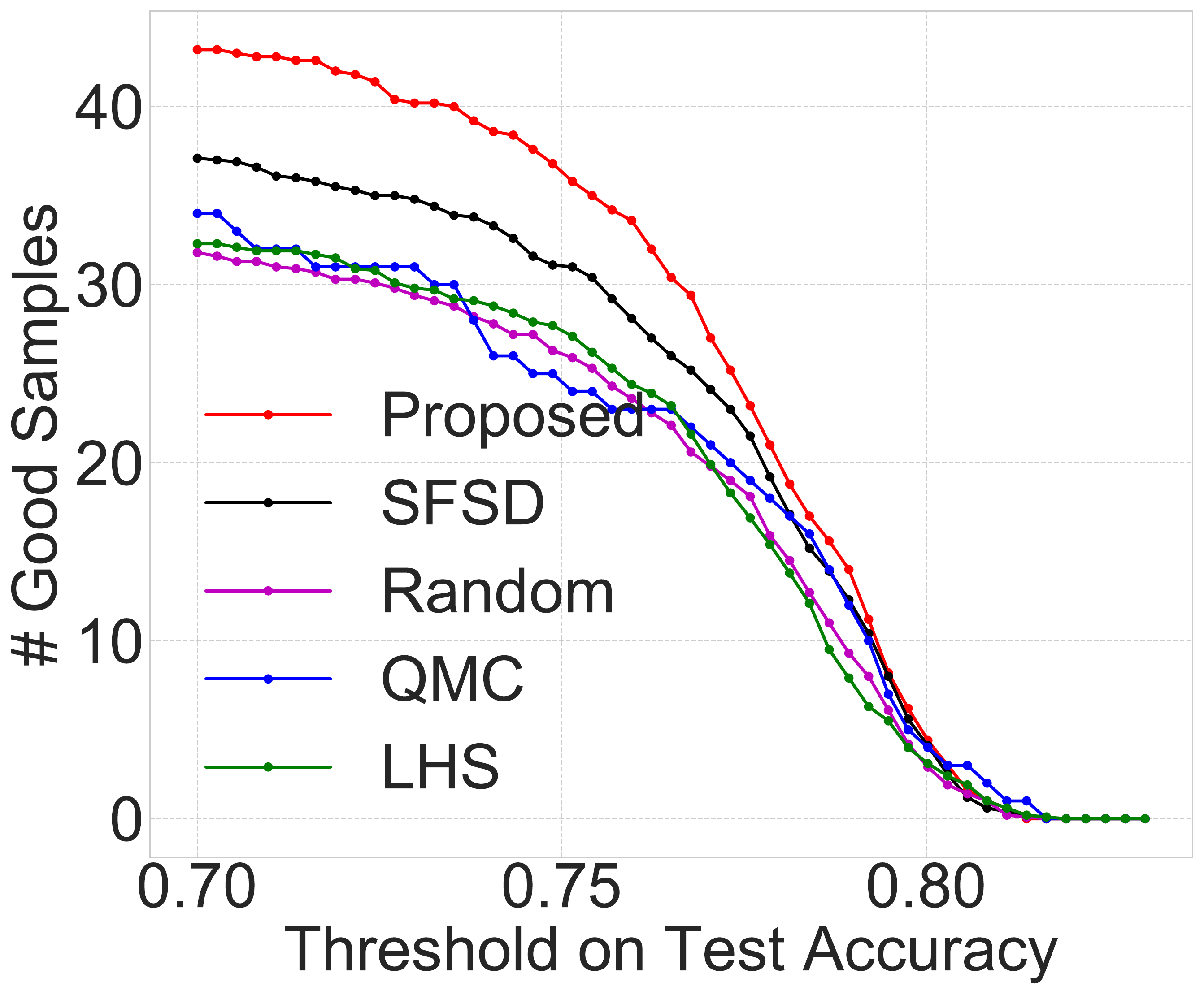} \label{fig:cif_bay_hopt_50_5D} 
	}    
	\caption{\textit{Hyper-parameter search to build CNNs for Cifar-10 image classification}: Precision metric obtained through blind exploration and Bayes-Opt with different initial exploratory samples.} 
	\label{fig:cifar_results}
\end{figure*}

\noindent \textbf{Sequential Sampling}: The success of Bayesian optimization relies on its ability to exploit uncertainties in the search space, to trade-off between exploration and exploitation. We argue that the choice of initial space-filling design can significantly impact the performance of sequential optimization in hyper-parameter search. For this experiment, we train a DNN architecture comprised of only dense layers: $\rm{FC[784] \to FC[512] \to FC[256] \to FC[64] \to FC[10] }$, with \rm{ReLU} activation and dropout after every layer. We used the same set of hyper-parameters as in the previous case, i.e., learning rate, momentum and dropout ratios. Experiments were conducted with an initial sampling budget of $N = 50$ samples and an additional $50$ samples from sequential sampling, in dimensions $3$, $4$ and $5$, respectively. Figure \ref{fig:MS_BO_results} and Table \ref{table:mnist} demonstrate the impact of different initial exploratory samples on the sequential optimization performance. The gains over discrepancy-based and random designs is even more significant in this case, thus emphasizing coverage as a desired characteristic of exploratory designs. The consistency of our approach in its performance across dimensions, evidences its robustness when compared to other widely adopted designs.

\vspace{-0.2cm} 
\subsection{Building CNN for Cifar-10 Image Classification}
In this final experiment, we consider a $5-d$ hyper-parameter search to train a CNN for classifying images from the CIFAR-10 dataset. The architecture used is as follows: $
 conv[3 \times 3 \times 32] \to conv[3 \times 3 \times 32] \to conv[3 \times 3 \times 64] \to conv[3 \times 3 \times 64] \to {\rm FC[512] \to FC[128] \to FC[10].
} $ with \rm{ReLU} activation, max-pooling, and batch normalization after every convolution layer. Dropouts are included after the $2^{nd}$ $conv$ layer, the $4^{th}$ $conv$ layer and the $1^{st}$ $FC$ layer. The set of $5$ hyper-parameters included learning rate, momentum and the $3$ dropout ratios. For blind exploration, we used sampling budgets of $N = 50$ and $N = 100$. In case of sequential sampling, experiments were conducted with an initial budget of $N = 50$ samples and an additional $50$ were sampled sequentially using Bayesian optimization. We report the mean and standard deviation of the best test accuracy achieved over $10$ realizations in Table~\ref{table:cifar} and the precision metric in Figure~\ref{fig:cifar_results}. In all the cases, the proposed method achieves the best expected generalization performance. Along with the other experiments, this observation clearly strengthens the premise that coverage-based designs are highly effective for hyper-parameter search when compared to existing random sampling and discrepancy-based designs.

\section{Conclusions}
We considered the problem of designing high quality exploratory samples. We introduced improved coverage Poisson disk sample designs using pair correlation function. We also proposed an approach to automatically determine the optimal parameters of the PDS designs. To generate these samples with high accuracy, we proposed an adaptive learning rate based gradient descent approach and showed that it significantly outperforms baseline methods. 
Finally, we evaluated the performance of PDS designs on predictive modeling and hyper-parameter search applications in both blind exploration and sequential search with Bayesian optimization. Experimental results 
show that the proposed PDS approach consistently outperforms state-of-the-art techniques, especially with low sampling budget.

\section{Acknowledgments}
This document was prepared as an account of work sponsored by an agency of the United States government. Neither the United States government nor Lawrence Livermore National Security, LLC, nor any of their employees makes any warranty, expressed or implied, or assumes any legal liability or responsibility for the accuracy, completeness, or usefulness of any information, apparatus, product, or process disclosed, or represents that its use would not infringe privately owned rights. Reference herein to any specific commercial product, process, or service by trade name, trademark, manufacturer, or otherwise does not necessarily constitute or imply its endorsement, recommendation, or favoring by the United States government or Lawrence Livermore National Security, LLC. The views and opinions of authors expressed herein do not necessarily state or reflect those of the United States government or Lawrence Livermore National Security, LLC, and shall not be used for advertising or product endorsement purposes.

\bibliographystyle{IEEEtran}
\bibliography{refs}
\end{document}